\documentclass[10pt,twocolumn,letterpaper]{article}

\usepackage{cvpr}              
\definecolor{cvprblue}{rgb}{0.21,0.49,0.74}
\usepackage[pagebackref,breaklinks,colorlinks,allcolors=cvprblue]{hyperref}
\usepackage{booktabs}   
\usepackage{multirow}   
\usepackage[table]{xcolor} 
\usepackage{threeparttable}

\usepackage{algpseudocode}
\usepackage{algorithm}
\usepackage{amsmath, amssymb}
\usepackage{bm}

\def\paperID{*****} 
\def\confName{CVPR}
\def\confYear{2026}

\title{ACE-Merging: Data-Free Model Merging with Adaptive Covariance Estimation}

\author{
Bo Xu$^{1}$ \quad
Haotian Wu$^{1}$ \quad
Hehai Lin$^{1}$ \quad
Weiquan Huang$^{1}$ \quad
Beier Zhu$^{2}$ \quad
Yao Shu$^{1}$ \quad
Chengwei Qin$^{1}$\thanks{Corresponding author}\\
$^{1}$The Hong Kong University of Science and Technology (Guangzhou)\\
$^{2}$University of Science and Technology of China\\
{\tt\small bx@hkust-gz.edu.cn \quad chengweiqin@hkust-gz.edu.cn}
}

\begin{document}

\newcommand{\acem}{\texttt{ACE-Merging}\xspace}

\maketitle
\begin{abstract}
Model merging aims to combine multiple task-specific experts into a single model, but inter-task interference often causes severe degradation, especially when the experts are trained on heterogeneous objectives. Existing data-free methods are practical, yet largely rely on parameter-space heuristics without explicitly modeling task statistics. In this paper, we show that under a local linear approximation, the input covariance of each task can be estimated directly from its fine-tuning update, providing a principled bridge between parameter changes and data geometry in the data-free setting. Based on this insight, we propose \acem, a closed-form model merging framework with adaptive covariance normalization, a collective structural prior, and spectral refinement. Across both vision and language benchmarks, \acem achieves strong overall performance among data-free baselines. In particular, it improves the average score on GPT-2 by more than 4 points over prior methods, while also delivering strong scalability and competitive efficiency. Our code is available at \url{https://github.com/unravel-xu/ACE-Merging/tree/main}.
\end{abstract}    
\section{Introduction}
\label{sec:intro}

Driven by advances in architectures such as the Transformer~\cite{vaswani2017attention}, the pre-training and fine-tuning paradigm~\cite{devlin2019bert} has produced a large collection of task-specialized expert models. In practice, however, deploying a separate expert for each task is often prohibitively costly, while retraining a unified multi-task model from scratch~\cite{caruana1997multitask} is frequently infeasible, especially when only fine-tuned checkpoints are available and the original training data are inaccessible. Model merging~\cite{ilharco2023editing,Wortsman2022ModelSA} provides an appealing alternative by combining multiple expert models directly in weight space, without requiring expensive joint retraining.

A central challenge in model merging is \emph{inter-task interference}. Existing methods can be broadly grouped into three categories. \emph{Data-dependent} approaches~\cite{matena2022fiser,jin2022regmean,nguyen2025regmean++} leverage input statistics to guide fusion, but require access to the original task data. \emph{Test-time adaptive} methods~\cite{yang2023adamerging,huang2024emr} adjust the merged model during inference, improving flexibility at the cost of additional deployment overhead. \emph{Data-free} methods are often the most practical because they operate solely on model parameters. Yet without direct access to task statistics, existing data-free approaches, from Task Arithmetic~\cite{ilharco2023editing} to more sophisticated variants~\cite{yadav2023ties,si2025nan,zheng2025free,yu2024DARE}, remain largely heuristic in how they mitigate interference across tasks.

In this work, we revisit data-free model merging from a statistical perspective. Rather than treating fine-tuning updates as unstructured parameter offsets, we show that under a local linear approximation they encode informative second-order structure related to the underlying task data. This observation provides a principled way to infer task geometry from weight updates alone, enabling a covariance-aware view of data-free model merging. Building on this perspective, we formulate merging as a closed-form optimization problem and instantiate the resulting framework as \acem.

Turning this formulation into a robust practical method requires addressing three challenges. First, second-order proxies derived from task vectors can differ substantially in scale across tasks, making direct aggregation unstable. Second, purely isotropic regularization is insufficient to capture structure shared among experts. Third, in highly heterogeneous settings, the resulting merge can exhibit an overly concentrated singular spectrum. \acem addresses these issues through three corresponding components: \emph{adaptive covariance normalization} to balance task scales, a \emph{collective structural prior} to capture shared geometry, and \emph{spectral refinement} to alleviate spectral concentration while preserving the dominant subspace. These components together make \acem effective and robust across diverse merging scenarios.

Extensive experiments on both vision and language benchmarks show that \acem consistently outperforms prior data-free baselines and achieves state-of-the-art performance in the data-free setting.

Our main contributions are summarized as follows:
\begin{itemize}
    \item \textbf{Theoretical Foundation.} We show that, under a local linear approximation, fine-tuning-induced weight displacements contain informative second-order structure related to task input statistics, providing a principled basis for covariance-aware data-free merging.
    \item \textbf{Unified Framework.} We propose \acem, a unified framework that combines adaptive covariance normalization, a collective structural prior, and spectral refinement to mitigate inter-task interference.
    \item \textbf{State-of-the-Art Performance.} We validate our method on diverse vision and language benchmarks, demonstrating clear improvements over prior data-free baselines with competitive efficiency and scalability.
\end{itemize}
\section{Related work}
\subsection{Data-dependent model merging}
Data-dependent model merging formulates the fusion as an optimization problem that relies on input-level statistics.
Representative approaches exploit feature covariances or Fisher information to weight parameters during aggregation~\cite{jin2022regmean,nguyen2025regmean++,matena2022fiser,lee2025dynamicfisher}. Although effective, their dependence on task data for estimating importance weights limits their applicability in fully data-free settings.




\subsection{Test-time adaptive model merging}
Test-time adaptive methods perform dynamic fusion during inference rather than constructing a single static model.  
They adjust merging coefficients or lightweight modules on the fly according to the current input or task context~\cite{yang2023adamerging,huang2024emr,bertolissi2025testtime,zheng2025free,qiu2025mingle}.  
Such approaches often minimize prediction uncertainty or employ routing mechanisms to specialize the merged model for unseen data.  
While these strategies can yield higher task-specific performance, they introduce additional inference overhead or rely on access to representative test data, limiting their practicality for static deployment.

\subsection{Data-free model merging}

In practical scenarios, access to task data is often limited due to privacy or scalability constraints, 
making data-free model merging the most general and desirable solution.
However, the absence of data also removes the statistical signals that reveal task relationships, 
forcing methods to rely on parameter-space heuristics to approximate the underlying data geometry.
Early data-free methods rely on mitigate interference through heuristic operations such as sign alignment or parameter pruning~\cite{yadav2023ties,deep2024della,yu2024DARE}. While efficient, these Euclidean-space operations fail to capture deeper structural misalignments between models. More recent work seeks a superior basis for merging, primarily through SVD-based decompositions that isolate shared and task-specific subspaces~\cite{gargiulo2025tsvm,marczak2025no,stoica2024knots,Panariello2025AccurateAE}. Concurrently, optimization-inspired methods like WUDI-Merging~\cite{cheng2025whoever} derive analytical solutions but often rely on iterative gradient descent for their implementation, sacrificing the stability of a true closed-form solution. In summary, data-free merging still lacks a unified and efficient closed-form formulation that captures inter-task structure. \acem fulfills this need through covariance estimation and spectrally adaptive fusion, achieving superior stability and efficiency over SVD- and gradient-based methods. (see Appendix \ref{app:complexity})

\section{Background and motivation}

In this section, we formalize the data-free model merging problem and derive the closed-form optimal solution under a local linearization assumption. This analysis provides the statistical motivation for our method.

\subsection{Preliminaries of model merging}

\noindent\textbf{Notation.}
Let $\mathcal{M}_0$ denote a pretrained model with layer weights $\{ W_0^{l} \}_{l \in [L]}$,\footnote{%
As our algorithm is applied independently to each layer, 
we omit the layer index $l$ in subsequent derivations for clarity.}
where $l$ indexes one of the $L$ layers.
Fine-tuning $\mathcal{M}_0$ on a downstream task $t$ yields a specialized model $\mathcal{M}_t$ with layer weights $\{W_t^l\}_{l \in [L]}$.
Following Task Arithmetic~\cite{ilharco2023editing}, we define the \emph{task vector} as the weight displacement
$\Delta W_t = W_t - W_0$, which captures the task-specific parameter update relative to the pretrained model.

\noindent\textbf{Model merging objective.}
Given $T$ fine-tuned expert models $\{\mathcal{M}_t\}_{t \in [T]}$ with parameters $\{ W_t \}_{t \in [T]}$,
our goal is to construct a unified merged model $\bar{\mathcal{M}}$ with parameters $\bar{W}$
that matches the outputs of all experts on their respective data distributions $\{\mathcal{D}_t\}$ as closely as possible.
Formally, we minimize the expected output discrepancy between the merged model and each expert:
\begin{equation}
    \min_{\bar{W}}
    \sum_{t \in [T]}
    \mathbb{E}_{x \sim \mathcal{D}_t}
    \left\|
        f(\bar{W}, x)
        -
        f(W_t, x)
    \right\|_2^2.
    \label{eq:merge_objective}
\end{equation}
Here, $f(W, x)$ denotes the layer-wise transformation induced by parameters $W$ on input feature $x$.

\subsection{Optimal merging under local linearization}
\label{subsec:optimal_linear_merging}

To obtain a tractable solution, we adopt the local linear approximation $f(W, x)\approx Wx$, which is commonly used in prior analyses~\cite{razzhigaev2024your}.
Under this approximation, Eq.~\eqref{eq:merge_objective} becomes
\begin{align}
    \mathcal{L}(\bar W)
    &=
    \sum_{t \in [T]}
    \mathbb{E}_{x \sim \mathcal{D}_t}
    \left\|
        (\bar W - W_t)x
    \right\|_2^2 \nonumber\\
    &=
    \sum_{t \in [T]}
    \operatorname{Tr}\left[
        (\bar W - W_t)\,
        \Sigma_t\,
        (\bar W - W_t)^\top
    \right],
    \label{eq:trace_form}
\end{align}
where $\Sigma_t = \mathbb{E}_{x \sim \mathcal{D}_t}[x x^\top]$ denotes the input second-moment matrix of task $t$.
Minimizing Eq.~\eqref{eq:trace_form} with respect to $\bar W$ yields the closed-form solution
\begin{equation}
    \bar W 
    =
    \Big(\sum_{t \in [T]} W_t \Sigma_t\Big)
    \Big(\sum_{t \in [T]} \Sigma_t\Big)^{-1},
    \label{eq:closed_form}
\end{equation}
assuming $\sum_{t \in [T]} \Sigma_t$ is invertible.
This expression shows that second-order task statistics determine the optimal merge under local linearization, and therefore provides a useful perspective for designing data-free merging rules through suitable second-order proxies.

\subsection{Baselines as second-order proxies}
\label{sec:revisiting_baselines}

Eq.~\eqref{eq:closed_form} also provides a convenient lens for interpreting representative data-free merging methods, including weight averaging~\cite{Wortsman2022ModelSA} and WUDI-Merging~\cite{cheng2025whoever}.

\paragraph{Weight averaging.}
The simplest baseline assumes uniform and isotropic task statistics, i.e., $\widehat{\Sigma}_t = kI$ for some constant $k>0$.
Substituting this assumption into Eq.~\eqref{eq:closed_form} yields
$\bar W = \frac{1}{T}\sum_{t=1}^{T} W_t$,
showing that weight averaging is optimal only under the restrictive condition that all tasks share identical isotropic second-order structure.

\paragraph{Task-specific second-order proxy from task vectors.}
A more expressive data-free strategy is to infer task-specific second-order structure from the task vector $\Delta W_t$.
A simple proxy for the input second-moment matrix is
$\widehat{\Sigma}_t \propto (\Delta W_t)^\top \Delta W_t$.
Although WUDI-Merging is derived from a different objective and optimized via gradient descent, it can be viewed through our framework as approximately adopting a norm-normalized version of this proxy:
$\widehat{\Sigma}_t \propto \|\Delta W_t\|_F^{-2} (\Delta W_t)^\top \Delta W_t$.
This perspective suggests that part of WUDI-Merging's effectiveness may come from using a more informative task-specific second-order proxy than uniform averaging. At the same time, its reliance on iterative optimization introduces additional computational overhead and hyperparameter sensitivity, motivating the closed-form estimator developed in the next section.
\section{Methodology}
\label{sec:method}

\subsection{Estimating input second-moment structure}

We first establish the connection between input second-order statistics and fine-tuning updates.

\paragraph{Theorem 1.}
\label{thm:cov_relation}
\textit{Under a local linearization with squared loss, the expected column Gram matrix of the per-sample gradient is aligned with the input second-moment matrix of task $t$ up to a task-dependent scalar:}
\begin{equation}
    \mathbb{E}_{(x,y)\sim \mathcal{D}_t}\!\left[g(x,y)^\top g(x,y)\right]
    \propto \Sigma_t,
    \quad
    \Sigma_t=\mathbb{E}_{x\sim\mathcal{D}_t}[xx^\top].
    \label{eq:matrix_cov_relation}
\end{equation}

\noindent\textit{Proof sketch.}
Under a squared-loss local linearization around $W_0$, the per-sample gradient can be written as
\[
g(x,y)=2e\,x^\top,\qquad e=W_0x-y .
\]
Therefore,
\[
g(x,y)^\top g(x,y)=4x\,e^\top e\,x^\top .
\]
Taking expectation over $(x,y)\sim\mathcal D_t$ gives
\[
\mathbb{E}_{(x,y)\sim\mathcal D_t}\!\left[g(x,y)^\top g(x,y)\right]
=
4\,\mathbb{E}_{(x,y)\sim\mathcal D_t}\!\left[(e^\top e)\,xx^\top\right].
\]
If the residual energy is approximately decoupled from the input direction in expectation, then
\[
\mathbb{E}_{(x,y)\sim\mathcal D_t}\!\left[(e^\top e)\,xx^\top\right]
\approx
\mathbb{E}_{(x,y)\sim\mathcal D_t}[e^\top e]\;
\mathbb{E}_{x\sim\mathcal D_t}[xx^\top],
\]
which yields
\[
\mathbb{E}_{(x,y)\sim\mathcal D_t}\!\left[g(x,y)^\top g(x,y)\right]
\propto
\mathbb{E}_{x\sim\mathcal D_t}[xx^\top]
=
\Sigma_t .
\]
Since the fine-tuning update $\Delta W_t$ accumulates such gradient contributions, its column Gram matrix $\Delta W_t^\top\Delta W_t$ provides a data-free second-order proxy for $\Sigma_t$ up to a task-dependent scale. A complete derivation is provided in Appendix \ref{app:theory}.
\hfill $\square$

Eq.~\eqref{eq:matrix_cov_relation} motivates a practical data-free estimator based on the centered column Gram matrix of the task vector. For each task, we define the centered task vector as
\begin{equation}
    \widetilde{W}_t
    =
    \Delta W_t - \mathbf{1}\mu_t^\top,
    \qquad
    \mu_t=\tfrac{1}{d_{\mathrm{out}}}\Delta W_t^\top \mathbf{1},
    \label{eq:centered_task_vector}
\end{equation}
where $\Delta W_t\in\mathbb R^{d_{\mathrm{out}}\times d_{\mathrm{in}}}$ and $\mathbf 1\in\mathbb R^{d_{\mathrm{out}}}$ is the all-ones vector.
We then use
\begin{equation}
    \widehat{\Sigma}_t
    \propto
    \widetilde{W}_t^\top \widetilde{W}_t
    =
    (\Delta W_t - \mathbf{1}\mu_t^\top)^\top
    (\Delta W_t - \mathbf{1}\mu_t^\top)
    \label{eq:empirical_cov_estimator}
\end{equation}
as the task-specific second-order proxy.
This centering removes the rank-one mean component across output dimensions, so that $\widehat{\Sigma}_t$ better captures task-specific second-order structure. The proportionality constant affects only the overall scale and therefore cancels in the subsequent closed-form merge.

Directly using this estimator, however, raises several practical issues, including scale imbalance across tasks, ill-conditioned aggregation, and instability under strong heterogeneity. To address these issues, we propose \acem, which consists of three components: (1) adaptive covariance normalization to balance task scales and stabilize inversion; (2) a collective structural prior to inject shared geometry across tasks; and (3) an optional spectral refinement step for highly heterogeneous settings. We describe each component below.

\subsection{Adaptive covariance normalization}
\label{sec:adaptive_cov_norm}

A key challenge in model merging is the large variation in update magnitudes across tasks.
When aggregating $\sum_{t} \widehat{\Sigma}_t$, high-energy tasks can dominate the sum, biasing the merged model toward their representations while suppressing lower-energy tasks.

\begin{figure*}[t]
\centering
\caption{
Comparison of inter-task heterogeneity ($\gamma$) across architectures.
ViT-B/16 shows relatively uniform scaling ($\gamma < 0.25$),
whereas RoBERTa-Base exhibits substantially stronger variation ($\gamma > 0.3$) across layers.
This contrast motivates the need for an adaptive scaling strategy.
}
\begin{subfigure}{0.49\linewidth}
    \includegraphics[width=\linewidth]{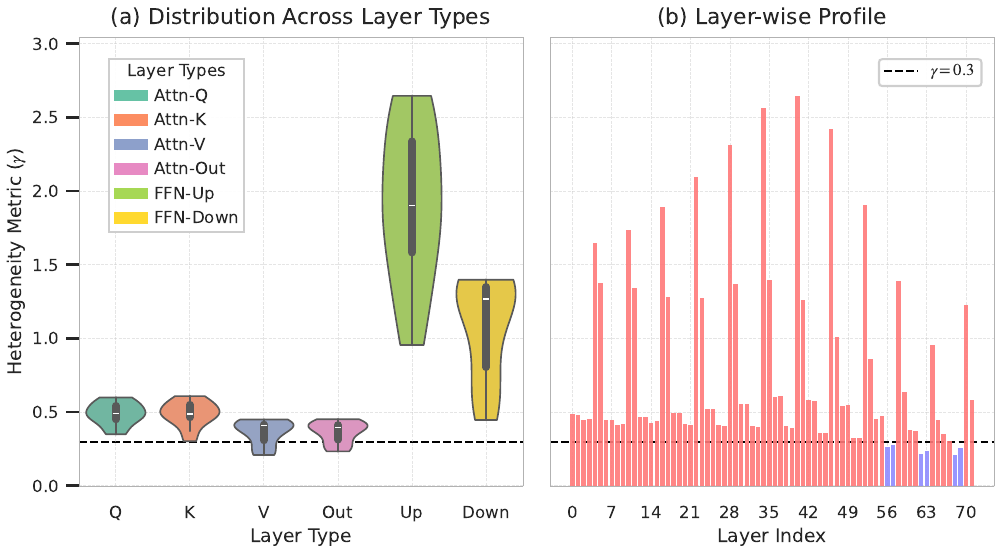}
    \caption{RoBERTa-Base on 8 tasks.}
    \label{fig:roberta_sub}
\end{subfigure}
\begin{subfigure}{0.49\linewidth}
    \includegraphics[width=\linewidth]{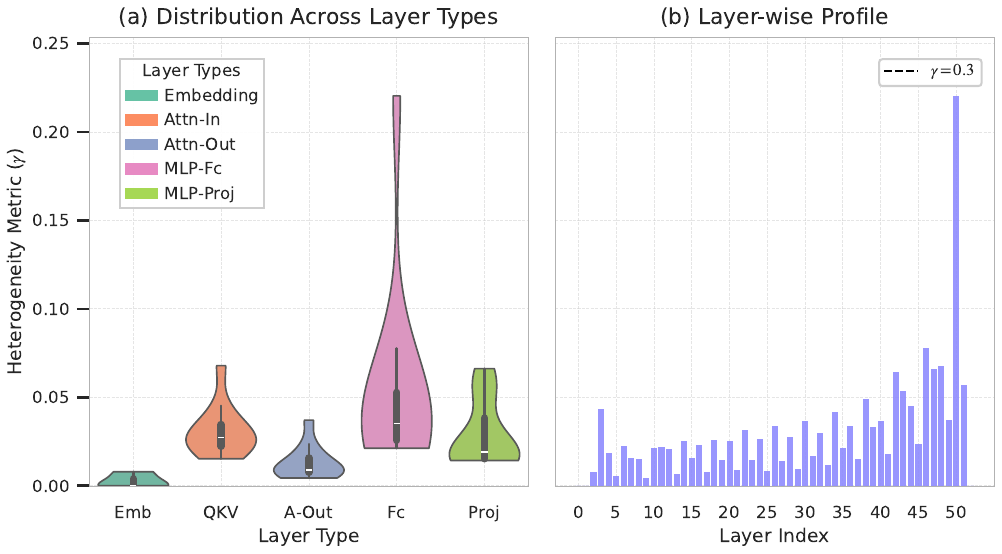}
    \caption{ViT-B/16 on 14 tasks.}
    \label{fig:vit_sub}
\end{subfigure}
\label{fig:gamma_across_models}
\end{figure*}

To mitigate this imbalance, we introduce an adaptive normalization strategy.
Rather than adjusting the regularization coefficient $\epsilon$ itself, we first decide whether the task-specific second-order estimators should be rescaled before aggregation.
This decision is guided by the heterogeneity statistic
\begin{equation}
    \gamma \;=\;
    \frac{\mathrm{Var}_{t}\!\left[\,\log \|\Delta W_t\|_{F}^{2}\,\right]}
         {\left(\mathbb{E}_{t}\!\left[\,\log \|\Delta W_t\|_{F}^{2}\,\right]\right)^{2}}.
    \label{eq:gamma_def}
\end{equation}
A larger $\gamma$ indicates stronger scale divergence across tasks.
As shown in Figure~\ref{fig:gamma_across_models}, this heterogeneity varies substantially across architectures, motivating an adaptive rather than uniform treatment.
Here, $\gamma$ is computed from the original task vectors $\Delta W_t$, since it is intended to measure task-wise energy disparity before any centering or normalization is applied.

When strong task heterogeneity is detected ($\gamma > \tau$),
we first normalize each estimator by its trace:
\begin{equation}
    \widehat{\Sigma}_{t,\text{scaled}} =
    \frac{\widehat{\Sigma}_t}{\operatorname{Tr}(\widehat{\Sigma}_t)}.
    \label{eq:adaptive_scaling}
\end{equation}
This removes task-wise energy disparities and places all tasks on a comparable scale.
We then apply Tikhonov regularization with an energy-adjusted coefficient:
\begin{equation}
    \widehat{\Sigma}_{t,\text{reg}}
    =
    \widehat{\Sigma}_{t,\text{scaled}}
    +
    \frac{\epsilon}{\operatorname{Tr}(\widehat{\Sigma}_t)} I.
    \label{eq:adaptive_regularization_policy}
\end{equation}

For relatively homogeneous task sets ($\gamma \le \tau$), we retain the original scale information and use the standard ridge-regularized estimator
\begin{equation}
    \widehat{\Sigma}_{t,\text{reg}}
    =
    \widehat{\Sigma}_t + \epsilon I.
\end{equation}
This branch-wise design preserves useful scale information when tasks are already well aligned, while enforcing scale balancing only when heterogeneity is pronounced.

\subsection{Collective structural prior}
\label{sec:collective_structural_prior}

\begin{figure*}[t]
    \centering
    \caption{
    Comparison between the preliminary closed-form solution $\bar W_{\mathrm{pre}}$
    and the final merged model $\bar W$.
    \textbf{Left:} The singular value spectrum of $\bar W_{\mathrm{pre}}$ is highly concentrated:
    the top 5\% singular values capture more than 99\% of the total energy, indicating severe spectral imbalance.
    In contrast, $\bar W$ exhibits a substantially flatter spectrum.
    \textbf{Right:} Despite this concentration, their leading singular vectors are nearly identical
    (cosine similarity $\approx 1$), showing that $\bar W_{\mathrm{pre}}$ already identifies the correct
    structural subspace. Spectral refinement preserves this subspace while restoring a more stable
    energy distribution.
    }\includegraphics[width=0.95\linewidth]{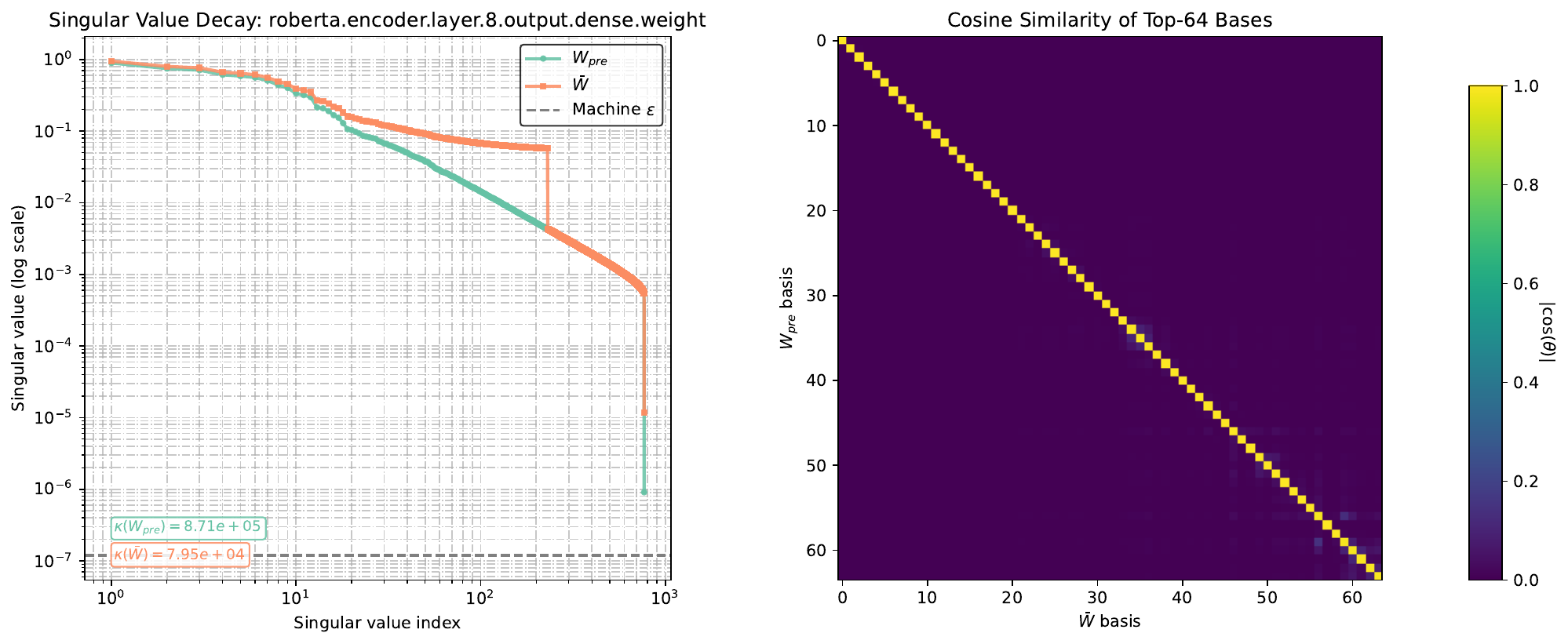}
    \label{fig:conditioning_and_projection}
\end{figure*}

While the strategy above balances task contributions and stabilizes matrix inversion, it remains isotropic:
the ridge term $\epsilon I$ treats all feature directions equally and does not exploit structure shared across tasks.

To incorporate such shared structure, we introduce a data-driven anisotropic regularizer, termed the \emph{Collective Structural Prior (CSP)}.
In our implementation, the CSP is constructed from the same pre-regularized matrices used before the final aggregation:
\[
\widetilde{\Sigma}_t =
\begin{cases}
\widehat{\Sigma}_{t,\mathrm{scaled}}, & \gamma > \tau,\\[0.5ex]
\widehat{\Sigma}_{t}, & \gamma \le \tau.
\end{cases}
\]
Let
\begin{equation}
c^\top
=
\frac{1}{d_{\mathrm{in}}}\,
\mathbf{1}^\top
\sum_{t \in [T]} \widetilde{\Sigma}_t,
\qquad
\mathbf{C}_{\mathrm{agg}} = \mathbf{1} c^\top,
\label{eq:c_agg}
\end{equation}
where $\mathbf{1}\in\mathbb{R}^{d_{\mathrm{in}}}$ is the all-ones vector.
Equivalently, $\mathbf{C}_{\mathrm{agg}}$ replicates the mean row of the aggregated task matrix across all rows, yielding a consensus-driven low-rank correction.
In implementation, this prior is realized by computing the mean row of $\sum_t \widetilde{\Sigma}_t$ and adding it to the system matrix through row-wise broadcasting.

In the heterogeneous regime ($\gamma>\tau$), we further rescale this prior by the average energy of the original task vectors:
\begin{equation}
\overline{\tau}
=
\frac{1}{T}\sum_{t\in[T]} \operatorname{Tr}(\Delta W_t^\top \Delta W_t)
=
\frac{1}{T}\sum_{t\in[T]} \|\Delta W_t\|_F^2,
\label{eq:c_agg_trace_mean}
\end{equation}
\begin{equation}
\mathbf{C}_{\mathrm{agg}}
\leftarrow
\frac{\mathbf{C}_{\mathrm{agg}}}{\overline{\tau}}.
\label{eq:c_agg_rescale}
\end{equation}
This keeps the scale of $\mathbf{C}_{\mathrm{agg}}$ compatible with the trace-normalized estimators used in the heterogeneous branch.

Using this prior, the preliminary closed-form merge becomes
\begin{equation}
\bar W_{\mathrm{pre}}
=
\Big(\sum_{t \in [T]} \widetilde{W}_t\,\widehat{\Sigma}_{t,\mathrm{reg}}\Big)
\Big(\sum_{t \in [T]} \widehat{\Sigma}_{t,\mathrm{reg}} + \mathbf{C}_{\mathrm{agg}}\Big)^{-1}.
\label{eq:pre_solution}
\end{equation}
That is, centering is used in the left multiplying task vectors that define the numerator, whereas the denominator is formed by the regularized second-order estimators together with the collective prior.

Unlike isotropic regularization, the CSP injects a shared structural bias into the merge, encouraging directions that are consistently emphasized across tasks. It therefore serves not only as a numerical stabilizer, but also as a simple geometry-aware prior.

\subsection{Spectral refinement}
\label{sec:spectral_correction}

Figure~\ref{fig:conditioning_and_projection} highlights a recurring phenomenon in highly heterogeneous settings: even after adaptive normalization and the collective prior $\mathbf{C}_{\mathrm{agg}}$, the preliminary solution $\bar W_{\mathrm{pre}}$ can remain spectrally ill-conditioned.
As shown in the left panel, the top 5\% singular values of $\bar W_{\mathrm{pre}}$ may account for more than 99\% of the total Frobenius energy, indicating extreme spectral concentration.
Such spike-shaped spectra make the merged model sensitive to perturbations and suggest that the closed-form solution may overemphasize only a few dominant directions.

At the same time, the right panel shows that the principal directions of $\bar W_{\mathrm{pre}}$ are nearly identical to those of the final merged model $\bar W$. This suggests that $\bar W_{\mathrm{pre}}$ already captures the correct structural subspace, and that its main deficiency lies in the singular value distribution rather than in the directions themselves. This motivates a refinement step that preserves the dominant subspace while restoring a more stable spectral profile.

Recall that $\widetilde{W}_t$ denotes the centered task vector used in covariance construction, and let
\begin{equation}
\widetilde{\Sigma}_t =
\begin{cases}
\widehat{\Sigma}_{t,\text{scaled}}, & \gamma > \tau,\\
\widehat{\Sigma}_t, & \gamma \le \tau.
\end{cases}
\end{equation}
We further define the mean regularized estimator as
\begin{equation}
\overline{\Sigma}_{\mathrm{reg}}
=
\tfrac{1}{T}\sum_{t\in[T]}\widehat{\Sigma}_{t,\text{reg}}.
\label{eq:sigma_reg_mean}
\end{equation}

We then compute the structural residual as
\begin{equation}
\Delta_{\mathrm{res}}
=
\sum_{t\in[T]}
\Delta W_t\bigl(\widetilde{\Sigma}_t-\overline{\Sigma}_{\mathrm{reg}}\bigr),
\label{eq:residual_correction_def}
\end{equation}
where the original uncentered task vectors $\Delta W_t$ are used in the left multiplication, rather than the centered vectors $\widetilde{W}_t$.
This residual acts as a first-order correction that reintroduces structural variation suppressed by aggregation.

We fuse this residual with the preliminary solution:
\[
\bar W_{\mathrm{fused}} = \bar W_{\mathrm{pre}} + \Delta_{\mathrm{res}}.
\]

Applying SVD to the fused matrix,
\[
\bar W_{\mathrm{fused}} = \mathbf{U}\mathbf{S}\mathbf{V}^\top,
\quad
\mathbf{S} = \operatorname{diag}(\sigma_1, \sigma_2, \dots),
\]
we rescale the top-$k$ directions by their mean singular value:
\begin{equation}
\Delta W_{\text{refine}}
= \sigma_{\text{iso}}\,\mathbf{U}_{:,1:k}\mathbf{V}_{:,1:k}^\top,
\quad
\sigma_{\text{iso}} = \tfrac{1}{k}\sum_{i=1}^{k}\sigma_i.
\label{eq:refinement_term_def}
\end{equation}

The final merged weight is then
\begin{equation}
\bar W = \bar W_{\mathrm{pre}} + \Delta W_{\mathrm{refine}}.
\label{eq:final_solution_with_refinement}
\end{equation}
This refinement is applied only when $\gamma > \tau$, i.e., under strong task heterogeneity.

\subsection{The ACE-Merging algorithm}

The complete \acem procedure is summarized in Algorithm~\ref{alg:ace_merging}. For each layer, it proceeds in three stages. It first estimates task heterogeneity from the original task vectors and determines whether scale normalization is needed. It then computes a robust closed-form merge using centered task vectors together with the regularized task-specific second-order estimators and the collective structural prior. Finally, when heterogeneity is strong, it applies spectral refinement by forming a residual with the original task vectors, improving the singular value distribution while preserving the dominant structural subspace.

\begin{algorithm}[ht!]
\caption{ACE-Merging (Layer-wise Procedure)}
\label{alg:ace_merging}
\small
\begin{algorithmic}[1]
\Require Task vectors $\{\Delta W_t\}_{t\in[T]}$, regularization strength $\epsilon$, heterogeneity threshold $\tau$, spectral rank fraction $k_{\mathrm{frac}}$
\Ensure Merged layer weights $\bar W$

\Statex \vspace{-2.4ex}
\Statex \textbf{\textit{Stage 1: Statistics Estimation \& Structural Prior}}
\State Compute heterogeneity score $\gamma$, centered task vectors $\{\widetilde W_t\}_{t\in[T]}$, raw covariance proxies $\{\widehat\Sigma_t\}_{t\in[T]}$ \hfill Eqs.~\eqref{eq:centered_task_vector}--\eqref{eq:gamma_def}
\State Construct collective structural prior $\mathbf C_{\mathrm{agg}}$ \hfill Eq.~\eqref{eq:c_agg}

\Statex \vspace{-2.4ex}
\Statex \textbf{\textit{Stage 2: Adaptive Covariance Normalization}}
\If{$\gamma > \tau$} \Comment{High heterogeneity regime}
    \State Apply Tikhonov regularization to obtain $\{\widehat\Sigma_{t,\mathrm{reg}}\}_{t\in[T]}$ \hfill Eqs.~\eqref{eq:adaptive_scaling},~\eqref{eq:adaptive_regularization_policy}
    \State Rescale collective prior: $\mathbf C_{\mathrm{agg}} \gets \mathbf C_{\mathrm{agg}} / \bar{\tau}$ \hfill Eq.~\eqref{eq:c_agg_rescale}
\Else \Comment{Low heterogeneity regime}
    \State Apply standard ridge regularization: $\widehat\Sigma_{t,\mathrm{reg}} \gets \widehat\Sigma_t + \epsilon I$ 
\EndIf

\Statex \vspace{-2.4ex}
\Statex \textbf{\textit{Stage 3: Closed-form Fusion \& Spectral Refinement}}
\State Compute preliminary merged weights $\bar W_{\mathrm{pre}}$ \hfill Eq.~\eqref{eq:pre_solution}

\If{$\gamma > \tau$} \Comment{Refine singular spectrum}
    \State Compute structural residual $\Delta_{\mathrm{res}}$ using original task vectors $\{\Delta W_t\}_{t\in[T]}$ \hfill Eq.~\eqref{eq:residual_correction_def}
    \State Extract low-rank refinement $\Delta W_{\mathrm{refine}}$ \hfill Eq.~\eqref{eq:refinement_term_def}
    \State \Return $\bar W \gets \bar W_{\mathrm{pre}} + \Delta W_{\mathrm{refine}}$ \hfill Eq.~\eqref{eq:final_solution_with_refinement}
\Else
    \State \Return $\bar W \gets \bar W_{\mathrm{pre}}$
\EndIf
\end{algorithmic}
\end{algorithm}

\begin{table*}[h]
\centering
\caption{Main results on vision benchmarks. We report the average absolute accuracy (\%) across three ViT backbones (ViT-B/32, ViT-B/16, and ViT-L/14) on task sets of increasing cardinality (8, 14, and 20). ``Zero-shot'' and ``Fine-tuned'' serve as reference lower and upper bounds, respectively. The best result among merging methods in each setting is highlighted in \textbf{bold}.}
\label{tab:main_results_vision}
\small
\begin{tabular}{l ccc ccc ccc}
\toprule
\multirow{2}{*}{\textbf{Method}}
& \multicolumn{3}{c}{\textbf{ViT-B/32}}
& \multicolumn{3}{c}{\textbf{ViT-B/16}}
& \multicolumn{3}{c}{\textbf{ViT-L/14}} \\
\cmidrule(lr){2-4} \cmidrule(lr){5-7} \cmidrule(lr){8-10}
& \textbf{8 tasks} & \textbf{14 tasks} & \textbf{20 tasks}
& \textbf{8 tasks} & \textbf{14 tasks} & \textbf{20 tasks}
& \textbf{8 tasks} & \textbf{14 tasks} & \textbf{20 tasks} \\
\midrule
\rowcolor{gray!15}
Zero-shot & 48.3 & 57.2 & 56.1 & 55.3 & 61.3 & 59.7 & 64.7 & 68.2 & 65.2 \\
\rowcolor{gray!15}
Fine-tuned & 92.8 & 90.9 & 91.3 & 94.6 & 92.8 & 93.2 & 95.8 & 94.3 & 94.7 \\
\midrule
Weight Averaging~\cite{Wortsman2022ModelSA}
& 66.3 & 64.3 & 61.0
& 72.2 & 69.5 & 65.3
& 79.6 & 76.7 & 71.6 \\
Task Arithmetic~\cite{ilharco2023editing}
& 70.8 & 65.3 & 60.5
& 75.4 & 70.5 & 65.8
& 84.9 & 79.4 & 74.0 \\
Consensus TA~\cite{Wang2024ConsensusTA}
& 75.0 & 70.4 & 65.4
& 79.4 & 74.4 & 69.8
& 86.3 & 82.2 & 79.0 \\
CART~\cite{Choi2024cart}
& 84.7 & 79.5 & 76.3
& 88.3 & 84.1 & 80.5
& 92.6 & 88.7 & 87.9 \\
TSV-M~\cite{gargiulo2025tsvm}
& 85.9 & 80.1 & 77.1
& 89.0 & 84.6 & 80.6
& 93.0 & 89.2 & 87.7 \\
Iso-C~\cite{marczak2025no}
& 86.3 & 80.3 & 75.5
& 90.6 & 84.8 & 79.6
& 94.2 & 89.3 & 87.6 \\
Iso-CTS~\cite{marczak2025no}
& 86.2 & 81.7 & 78.1
& \textbf{91.1} & \textbf{86.4} & \textbf{82.4}
& \textbf{94.7} & 91.0 & \textbf{90.1} \\
\midrule
\rowcolor{blue!10}
\textbf{ACE-Merging (Ours)}
& \textbf{87.9} & \textbf{82.3} & \textbf{78.8}
& 90.6 & 86.1 & 82.1
& 94.6 & \textbf{91.1} & 89.5 \\
\bottomrule
\end{tabular}
\end{table*}

\begin{table*}[h]
\centering
\caption{Performance on the GLUE benchmark for different merging methods applied to GPT-2. All metrics are reported in their standard evaluation form. ``Individual'' denotes the upper bound obtained by single-task fine-tuned models.}
\label{tab:gpt2_results}
\small
\setlength{\tabcolsep}{5pt}
\begin{tabular}{l ccccccc | c}
\toprule
\textbf{Method} & \textbf{CoLA} & \textbf{MNLI} & \textbf{MRPC} & \textbf{QNLI} & \textbf{QQP} & \textbf{RTE} & \textbf{SST-2} & \textbf{Avg.} \\
\midrule
\rowcolor{gray!15}
Individual & 76.8 & 82.1 & 80.4 & 88.3 & 89.6 & 65.3 & 91.2 & 82.0 \\
\midrule
Weight Averaging~\cite{Wortsman2022ModelSA} & 55.0 & 55.1 & 51.0 & 57.6 & 76.7 & 44.8 & 52.5 & 56.1 \\
Fisher Merging~\cite{matena2022fiser} & 54.8 & 58.0 & 39.5 & 63.3 & 81.5 & 49.1 & 64.7 & 58.7 \\
Iso-CTS~\cite{marczak2025no} & 58.4 & 56.1 & 43.4 & 56.0 & 79.7 & 51.3 & 67.0 & 58.82 \\
Iso-C~\cite{marczak2025no} & 65.7 & 59.0 & 48.8 & 57.8 & 83.6 & 52.4 & 66.4 & 61.95 \\
RegMean~\cite{jin2022regmean} & 61.7 & 70.4 & 65.4 & 69.7 & 78.8 & 56.0 & 79.7 & 68.8 \\
Task Arithmetic~\cite{ilharco2023editing} & 68.7 & 68.6 & 69.6 & 70.5 & 81.8 & 47.3 & 83.6 & 70.0 \\
Ties-Merging~\cite{yadav2023ties} & 68.4 & 71.4 & 68.4 & 69.6 & 82.4 & 47.7 & 81.8 & 70.0 \\
TSV-M~\cite{gargiulo2025tsvm} & 65.6 & \textbf{75.4} & 58.6 & 64.4 & \textbf{86.2} & 55.6 & \textbf{85.7} & 70.2 \\
\midrule
\rowcolor{blue!10}
\textbf{ACE-Merging (Ours)} & \textbf{70.3} & 69.9 & \textbf{71.8} & \textbf{76.7} & 79.0 & \textbf{62.5} & \textbf{88.5} & \textbf{74.1} \\
\bottomrule
\end{tabular}
\end{table*}

\begin{table*}[h]
\centering
\caption{Normalized performance on the GLUE benchmark for RoBERTa-Base and RoBERTa-Large. The average score is reported, and the best results in each column are shown in \textbf{bold}.}
\label{tab:roberta_results}
\small
\setlength{\tabcolsep}{5pt}
\begin{tabular}{ll cccccccc | c}
\toprule
\textbf{Model} & \textbf{Method} & \textbf{CoLA} & \textbf{SST-2} & \textbf{MRPC} & \textbf{STS-B} & \textbf{QQP} & \textbf{QNLI} & \textbf{MNLI} & \textbf{RTE} & \textbf{Avg.} \\
\midrule
\multirow{8}{*}{RoBERTa-Base}
& Pre-trained & 0.0 & 53.8 & 85.0 & 4.0 & 37.5 & 53.1 & 37.1 & 71.2 & 41.7 \\
& Individual & 100.0 & 100.0 & 100.0 & 100.0 & 100.0 & 100.0 & 100.0 & 100.0 & 100.0 \\
\cmidrule{2-11}
& Weight Averaging~\cite{Wortsman2022ModelSA} & 0.0 & 59.2 & 85.8 & 47.0 & 45.4 & 63.9 & 48.0 & 71.2 & 52.6 \\
& Task Arithmetic~\cite{ilharco2023editing} & 8.4 & 88.3 & 89.6 & 32.8 & 82.0 & 85.4 & 75.5 & 80.4 & 67.8 \\
& Ties-Merging~\cite{yadav2023ties} & 31.8 & 88.9 & 86.2 & 10.9 & 61.1 & 85.9 & 83.0 & 69.6 & 64.7 \\
& Task Arithmetic (w/ DARE)~\cite{yu2024DARE} & 0.0 & 88.1 & 86.6 & 30.2 & 84.3 & 79.1 & 64.0 & 77.2 & 63.7 \\
& Ties-Merging (w/ DARE)~\cite{yu2024DARE} & 11.8 & 95.5 & 85.8 & 9.4 & 86.8 & 88.7 & 83.1 & 65.6 & 65.6 \\
& WUDI-Merging~\cite{cheng2025whoever} & 81.8 & \textbf{98.3} & 78.7 & 60.5 & \textbf{92.7} & \textbf{90.5} & \textbf{93.3} & 86.4 & 85.3 \\
\midrule
\rowcolor{blue!10}
& \textbf{ACE-Merging (Ours)} & \textbf{97.3} & 97.6 & \textbf{93.9} & \textbf{71.1} & 88.3 & 88.2 & 91.5 & \textbf{95.1} & \textbf{90.4} \\
\midrule
\multirow{8}{*}{RoBERTa-Large}
& Pre-trained & 0.0 & 51.5 & 40.9 & 20.9 & 36.4 & 56.0 & 37.6 & 62.4 & 38.2 \\
& Individual & 100.0 & 100.0 & 100.0 & 100.0 & 100.0 & 100.0 & 100.0 & 100.0 & 100.0 \\
\cmidrule{2-11}
& Weight Averaging~\cite{Wortsman2022ModelSA} & 7.4 & 55.1 & 84.2 & 46.3 & 56.7 & 73.8 & 35.8 & 66.7 & 53.3 \\
& Task Arithmetic~\cite{ilharco2023editing} & 7.4 & 86.1 & 86.8 & 78.0 & 90.7 & 77.0 & 73.3 & 67.6 & 70.9 \\
& Ties-Merging~\cite{yadav2023ties} & 42.7 & 78.1 & 85.2 & 51.7 & 89.9 & 81.9 & 79.7 & 70.0 & 72.4 \\
& Task Arithmetic (w/ DARE)~\cite{yu2024DARE} & 4.1 & 85.2 & 85.8 & 71.6 & 91.3 & 85.6 & 75.2 & 68.1 & 70.9 \\
& Ties-Merging (w/ DARE)~\cite{yu2024DARE} & 2.9 & 90.4 & 86.8 & 75.4 & 92.4 & 86.4 & 79.0 & 69.1 & 72.8 \\
& WUDI-Merging~\cite{cheng2025whoever} & 82.2 & 98.7 & 87.3 & 81.4 & 94.6 & 96.6 & \textbf{93.4} & 77.1 & 88.8 \\
\midrule
\rowcolor{blue!10}
& \textbf{ACE-Merging (Ours)} & \textbf{87.1} & \textbf{99.2} & \textbf{92.3} & \textbf{82.1} & \textbf{95.6} & \textbf{97.9} & 92.6 & \textbf{86.7} & \textbf{91.7} \\
\bottomrule
\end{tabular}
\end{table*}
\section{Experiments}
\label{sec:experiments}

\subsection{Experimental setup}

\paragraph{Datasets.}
For vision tasks, we evaluate on three vision benchmarks with cardinalities of 8, 14, and 20.
These benchmarks are constructed by progressively extending the task selection protocol of~\cite{gargiulo2025tsvm}.
The 8-task benchmark serves as our base setting and includes Cars~\cite{krause2013cars}, DTD~\cite{cimpoi2014DTD}, EuroSAT~\cite{helber2019eurosat}, GTSRB~\cite{stallkamp2011gtsrb}, MNIST~\cite{lecun2002mnist}, RESISC45~\cite{cheng2017resisc45}, SUN397~\cite{xiao2016sun397}, and SVHN~\cite{netzer2011svhn}.
The 14-task benchmark further adds CIFAR100~\cite{krizhevsky2009cifar100}, STL10~\cite{coates2011stl10}, Flowers102~\cite{nilsback2008flowers102}, OxfordIIITPet~\cite{parkhi2012OxfordIIITPet}, PCAM~\cite{veeling2018pcam}, and FER2013~\cite{goodfellow2013fer2013}.
The 20-task benchmark further adds EMNIST~\cite{cohen2017emnist}, CIFAR10~\cite{krizhevsky2009cifar100}, Food101~\cite{bossard2014food}, FashionMNIST~\cite{xiao2017fashionmnist}, RenderedSST2~\cite{socher2013renderedSST2}, and KMNIST~\cite{clanuwat2018kmnist}.

For language tasks, we evaluate on the GLUE benchmark~\cite{wang2018glue} and report results on its eight core tasks: CoLA, SST-2, MRPC, STS-B, QQP, QNLI, MNLI, and RTE.

\paragraph{Models.}
For vision tasks, we adopt CLIP-derived Vision Transformers (ViTs)~\cite{dosovitskiy2020vit,radford2021clip}, including ViT-B/32, ViT-B/16, and ViT-L/14.
The corresponding fine-tuned checkpoints are taken from the TSV-M suite~\cite{gargiulo2025tsvm}.

For language tasks, we consider both the RoBERTa~\cite{Liu2019roberta} and GPT-2~\cite{radford2019gpt2} model families, conducting experiments on RoBERTa-Base, RoBERTa-Large, and GPT-2.
RoBERTa checkpoints are sourced from the Twin-Merging benchmark~\cite{lu2024twin}, while GPT-2 checkpoints are obtained from FusionBench~\cite{tang2024fusionbench}.
We also evaluate on Llama-3B-Instruct~\cite{Touvron2023LLaMAOA}, with results reported in Appendix \ref{app:vit_scaling}.

\paragraph{Baselines.}
We compare against representative recent merging methods commonly used in prior work, including Weight Averaging~\cite{Wortsman2022ModelSA}, Task Arithmetic~\cite{ilharco2023editing}, Ties-Merging~\cite{yadav2023ties}, CART~\cite{Choi2024cart}, TSV-M~\cite{gargiulo2025tsvm}, WUDI-Merging~\cite{cheng2025whoever}, and data-dependent baselines such as Fisher Merging~\cite{matena2022fiser} and RegMean~\cite{jin2022regmean} when available.

\paragraph{Implementation details.}
Unless otherwise stated, we use the same hyperparameter setting across experiments.
The heterogeneity threshold is fixed to $\tau = 0.3$, and the spectral rank fraction is set to $k_{\mathrm{frac}} = 0.3$, corresponding to
$k = k_{\mathrm{frac}} \times \min(d_{\mathrm{in}}, d_{\mathrm{out}})$ principal components in the refinement step.
The regularization strength $\epsilon$ is chosen by model family: $4 \times 10^{-2}$ for GPT-2, $2 \times 10^{-4}$ for RoBERTa-Base, and $1 \times 10^{-5}$ for all other models.


\subsection{Main results}
\label{subsec:main_results}

We evaluate \acem on a broad suite of vision and language benchmarks.
Tables~\ref{tab:main_results_vision}, \ref{tab:gpt2_results}, and \ref{tab:roberta_results} summarize the main results.
Across these benchmarks, \acem remains highly competitive and achieves the strongest overall performance among data-free baselines.

\paragraph{Vision task performance.}
Table~\ref{tab:main_results_vision} reports the average accuracy on the 8-, 14-, and 20-task vision benchmarks. \acem remains highly competitive across all nine settings and achieves the strongest overall average performance. While \textsc{Iso-CTS} performs better in several individual configurations, \acem attains the top result on all ViT-B/32 settings and on the 14-task ViT-L/14 benchmark. These results highlight \acem as a robust and scalable solution across varying model capacities.

\paragraph{Language task performance.}
On GPT-2 (Table~\ref{tab:gpt2_results}), \acem achieves an average score of 74.1, substantially outperforming all baselines, including Ties-Merging (70.0) and TSV-M (70.2), by more than 4 absolute points.

On RoBERTa (Table~\ref{tab:roberta_results}), \acem also delivers consistent gains across both model scales.
For RoBERTa-Base, it improves the normalized average score from 85.3 for WUDI-Merging to 90.4.
For RoBERTa-Large, it further reaches 91.7, outperforming WUDI-Merging by nearly 3 points.
Taken together, the results across vision and language benchmarks suggest that the advantages of \acem become especially pronounced when task heterogeneity is substantial, which is consistent with the design of our method.

\subsection{Ablation study}
\label{subsec:ablation}

We ablate the three components of \acem on RoBERTa-Large, GPT-2, and ViT-B/16 (8 tasks), starting from a basic closed-form baseline (E1) and progressively adding adaptive covariance normalization (E2), the collective structural prior (E3), and spectral refinement (E4). Results are reported in Table~\ref{tab:ablation}.

\paragraph{Component analysis.}
On RoBERTa-Large and GPT-2, adaptive normalization provides the largest gain, highlighting the importance of handling cross-task scale heterogeneity. The collective structural prior yields a smaller but complementary effect, and the full model with spectral refinement achieves the best performance on both language benchmarks. On ViT-B/16, the heterogeneity statistic remains below the threshold $\tau=0.3$ across all layers, so \acem correctly skips the adaptive and refinement branches, which explains why E2 matches E1 and E4 matches E3. The improvement from E2 to E3 further shows that the collective structural prior remains useful even in relatively homogeneous settings.

\paragraph{Sensitivity analysis.}
Table~\ref{tab:sensitivity} shows that \acem is robust to moderate variations in both $\tau$ and $k_{\mathrm{frac}}$. Performance remains largely stable over $\tau \in [0.1, 0.3]$ and $k_{\mathrm{frac}} \in [0.1, 0.5]$. In contrast, the regularization strength $\epsilon$ more directly controls the balance between numerical stability and fidelity, suggesting that a principled strategy for choosing $\epsilon$ could further improve the method.

\begin{table}[t]
\centering
\small 
\renewcommand{\arraystretch}{0.95} 

\caption{Ablation study of ACE-Merging components. We report the average score (\%) on RoBERTa-Large (GLUE), GPT-2 (GLUE), and ViT-B/16 (8 tasks). The best configuration for each model is highlighted in \textbf{bold}.}
\label{tab:ablation}
\setlength{\tabcolsep}{2pt}
\begin{tabular}{l l c c c}
\toprule
\# & Method Configuration & RoBERTa-L & GPT-2 & ViT-B/16 \\
\midrule
E1 & Basic closed-form & 80.05 & 68.72 & 89.91 \\
E2 & + Adaptive normalization & 88.04 & 71.50 & 89.91 \\
E3 & + Collective structural prior & 86.79 & 71.51 & \textbf{90.60} \\
\rowcolor{gray!15}
E4 & + Spectral refinement & \textbf{91.68} & \textbf{74.09} & \textbf{90.60} \\
\bottomrule
\end{tabular}

\vspace{2mm} 

\caption{Sensitivity to the heterogeneity threshold $\tau$ (fixed $k_{\mathrm{frac}}\!=\!0.3$) and spectral rank fraction $k_{\mathrm{frac}}$ (fixed $\tau\!=\!0.3$). Average accuracy (\%) on RoBERTa-Large and GPT-2.}
\label{tab:sensitivity}
\setlength{\tabcolsep}{8pt}
\begin{tabular}{c | cc | cc}
\toprule
\multirow{2}{*}{\textbf{Value}} & \multicolumn{2}{c|}{\textbf{RoBERTa-Large}} & \multicolumn{2}{c}{\textbf{GPT-2}} \\
\cmidrule(lr){2-3} \cmidrule(lr){4-5}
 & $\tau$ & $k_{\mathrm{frac}}$ & $\tau$ & $k_{\mathrm{frac}}$ \\
\midrule
0.1 & \textbf{92.22} & 90.71 & 73.96 & \textbf{74.17} \\
0.2 & 92.21 & 91.55 & 73.96 & 73.88 \\
0.3 & 91.68 & \textbf{91.68} & \textbf{74.09} & 74.09 \\
0.4 & 89.07 & 91.61 & 71.74 & 74.06 \\
0.5 & 88.09 & 91.45 & 69.35 & 73.78 \\
\bottomrule
\end{tabular}

\vspace{-5mm}
\end{table}
\section{Conclusion}
\label{sec:conclusion}

In this work, we study data-free model merging under substantial inter-task heterogeneity. We propose \acem, a unified framework that combines adaptive normalization, a collective structural prior, and spectral refinement to integrate shared knowledge while preserving informative task-specific structure. Extensive experiments on diverse vision and language benchmarks show that \acem remains highly competitive across settings and achieves strong overall performance among data-free baselines. Its advantages become particularly evident as task heterogeneity increases, highlighting the robustness and scalability of the proposed approach. An interesting direction for future work is to make the regularization strength $\epsilon$ adaptive rather than treating it as a fixed global constant. We hope that \acem serves both as a practical method and as a useful statistical perspective for future research on scalable, data-free model merging.
{
    \small
    \bibliographystyle{unsrt}
    \bibliography{references}
}
\clearpage
\onecolumn
\appendix
\clearpage
\setcounter{page}{1}

\section{Theoretical Derivations}
\label{app:theory}

\subsection{Derivation of the Quadratic Objective}

We derive the quadratic objective in Eq.~\eqref{eq:trace_form} from the layer-wise discrepancy objective in Eq.~\eqref{eq:merge_objective}

Recall the merging objective:
\begin{equation}
\min_{\bar W}
\sum_{t\in[T]}
\mathbb{E}_{x\sim \mathcal D_t}
\bigl\|
f(\bar W,x)-f(W_t,x)
\bigr\|_2^2 .
\label{eq:a1_obj_start}
\end{equation}

Under the local linear approximation \(f(W,x)\approx Wx\), the objective becomes
\begin{equation}
\sum_{t\in[T]}
\mathbb{E}_{x\sim \mathcal D_t}
\bigl\|
(\bar W-W_t)x
\bigr\|_2^2 .
\label{eq:a1_linearized}
\end{equation}

For any matrix \(A\) and vector \(x\), we have
\[
\|Ax\|_2^2=x^\top A^\top A x
=\mathrm{Tr}(Axx^\top A^\top).
\]
Applying this identity with \(A=\bar W-W_t\), we obtain
\begin{align}
\mathbb{E}_{x\sim\mathcal D_t}
\|(\bar W-W_t)x\|_2^2
&=
\mathbb{E}_{x\sim\mathcal D_t}
\mathrm{Tr}\!\left((\bar W-W_t)xx^\top(\bar W-W_t)^\top\right) \\
&=
\mathrm{Tr}\!\left(
(\bar W-W_t)\,
\mathbb{E}_{x\sim\mathcal D_t}[xx^\top]\,
(\bar W-W_t)^\top
\right).
\end{align}

Defining
\[
\Sigma_t:=\mathbb{E}_{x\sim\mathcal D_t}[xx^\top],
\]
we arrive at
\begin{equation}
L(\bar W)
=
\sum_{t\in[T]}
\mathrm{Tr}\!\left(
(\bar W-W_t)\Sigma_t(\bar W-W_t)^\top
\right),
\label{eq:a1_quadratic}
\end{equation}
which is exactly Eq.~\eqref{eq:trace_form} in the main paper.

\subsection{Derivation of the Closed-Form Solution}
\label{app:derivation_closed_form}

In this section, we derive the closed-form solution for the optimal merged weights $\bar{W}$ by minimizing the quadratic objective function from Eq.~\eqref{eq:trace_form}.

To find the minimum, we compute the gradient of $\mathcal{L}(\bar{W})$ with respect to the matrix $\bar{W}$ and set it to zero. First, let's expand the term inside the trace:
\begin{align}
    (\bar{W} - W_t) \Sigma_t (\bar{W} - W_t)^\top &= (\bar{W} - W_t) \Sigma_t (\bar{W}^\top - W_t^\top) \nonumber \\
    &= \bar{W}\Sigma_t\bar{W}^\top - \bar{W}\Sigma_t W_t^\top - W_t\Sigma_t\bar{W}^\top + W_t\Sigma_t W_t^\top.
\end{align}
Using the linearity of the trace and the sum, we can differentiate the loss term by term with respect to $\bar{W}$. We use the following standard matrix calculus identities for the gradient of a trace:
\begin{itemize}
    \item $\nabla_X \operatorname{Tr}(XAX^\top) = X(A+A^\top)$
    \item $\nabla_X \operatorname{Tr}(XB) = B^\top$
    \item $\nabla_X \operatorname{Tr}(AX^\top) = A$
\end{itemize}
Given that the covariance matrix $\Sigma_t$ is symmetric (i.e., $\Sigma_t = \Sigma_t^\top$), the first identity simplifies to $\nabla_X \operatorname{Tr}(X\Sigma_t X^\top) = 2X\Sigma_t$. Applying these rules, the gradient of the loss function is:
\begin{align}
    \nabla_{\bar{W}} \mathcal{L}(\bar{W}) &= \nabla_{\bar{W}} \sum_{t \in [T]} \operatorname{Tr}\left[ \bar{W}\Sigma_t\bar{W}^\top - \bar{W}\Sigma_t W_t^\top - W_t\Sigma_t\bar{W}^\top + W_t\Sigma_t W_t^\top \right] \nonumber \\
    &= \sum_{t \in [T]} \nabla_{\bar{W}} \operatorname{Tr}\left[ \bar{W}\Sigma_t\bar{W}^\top - \bar{W}\Sigma_t W_t^\top - W_t\Sigma_t\bar{W}^\top \right] \nonumber \\
    &= \sum_{t \in [T]} \left( 2\bar{W}\Sigma_t - (\Sigma_t W_t^\top)^\top - W_t\Sigma_t \right) \nonumber \\
    &= \sum_{t \in [T]} \left( 2\bar{W}\Sigma_t - W_t\Sigma_t^\top - W_t\Sigma_t \right) \nonumber \\
    &= \sum_{t \in [T]} \left( 2\bar{W}\Sigma_t - 2W_t\Sigma_t \right) \quad (\text{since } \Sigma_t = \Sigma_t^\top) \nonumber \\
    &= 2 \sum_{t \in [T]} (\bar{W}\Sigma_t - W_t\Sigma_t).
\end{align}
Setting the gradient to zero to find the minimum:
\begin{align}
    & 2 \sum_{t \in [T]} (\bar{W}\Sigma_t - W_t\Sigma_t) = 0 \nonumber \\
    & \sum_{t \in [T]} \bar{W}\Sigma_t = \sum_{t \in [T]} W_t\Sigma_t .
\end{align}
Since $\bar{W}$ is common to all terms in the left-hand summation, we can factor it out:
\begin{equation}
    \bar{W} \left( \sum_{t \in [T]} \Sigma_t \right) = \sum_{t \in [T]} W_t\Sigma_t .
\end{equation}
Finally, to isolate $\bar{W}$, we right-multiply both sides by the inverse of the aggregated covariance matrix:
\begin{equation}
    \bar{W} = \left( \sum_{t \in [T]} W_t\Sigma_t \right) \left( \sum_{t \in [T]} \Sigma_t \right)^{-1}.
\end{equation}
This gives the closed-form solution for the optimal merged weights,
which matches Eq.~\eqref{eq:closed_form} in the main paper.

\subsection{Proof of Theorem 1}
\label{app:proof_theorem_1}

We provide a detailed derivation showing that the second-order structure
of fine-tuning updates is governed by the input second-moment matrix.

Let $W_0 \in \mathbb{R}^{d_{\mathrm{out}}\times d_{\mathrm{in}}}$ denote
the pretrained weights, and let $W_t$ be the task-specific weights obtained
after fine-tuning on dataset
$\mathcal{D}_t = \{(x_i,y_i)\}_{i=1}^{N_t}$.
Define the weight displacement
\[
\Delta W_t = W_t - W_0.
\]

\paragraph{Assumptions.}
We adopt the following standard assumptions.

\begin{itemize}
    \item[(A1)] \textbf{Local linearization with squared loss.}
    The loss is $\mathcal{L}(x,y;W)=\|Wx-y\|_2^2$, and its gradient at $W_0$ is
    \[
    \nabla_W \mathcal{L}(x,y;W_0)
    = 2(W_0x-y)x^\top
    = 2e\,x^\top,
    \]
    where $e = W_0x - y \in \mathbb{R}^{d_{\mathrm{out}}}$.

    \item[(A2)] \textbf{i.i.d.\ sampling.}
    Samples $(x_i,y_i)$ are drawn i.i.d.\ from $\mathcal{D}_t$.

    \item[(A3)] \textbf{Approximately zero-mean first-order update.}
    \[
    \mathbb{E}_{(x,y)\sim\mathcal{D}_t}[e\,x^\top] \approx 0.
    \]

    \item[(A4)] \textbf{Residual-input decoupling.}
    \[
    \mathbb{E}_{(x,y)\sim\mathcal{D}_t}\!\left[(e^\top e)\,xx^\top\right]
    \approx
    \mathbb{E}_{(x,y)\sim\mathcal{D}_t}[e^\top e]\,
    \mathbb{E}_{x\sim\mathcal{D}_t}[xx^\top].
    \]
\end{itemize}

Let
\[
\Sigma_t = \mathbb{E}_{x\sim\mathcal{D}_t}[xx^\top]
\]
denote the task-specific input second-moment matrix.

\begin{figure*}[htbp]
    \centering

    \begin{subfigure}[b]{0.33\textwidth}
        \centering
        \includegraphics[width=\linewidth]{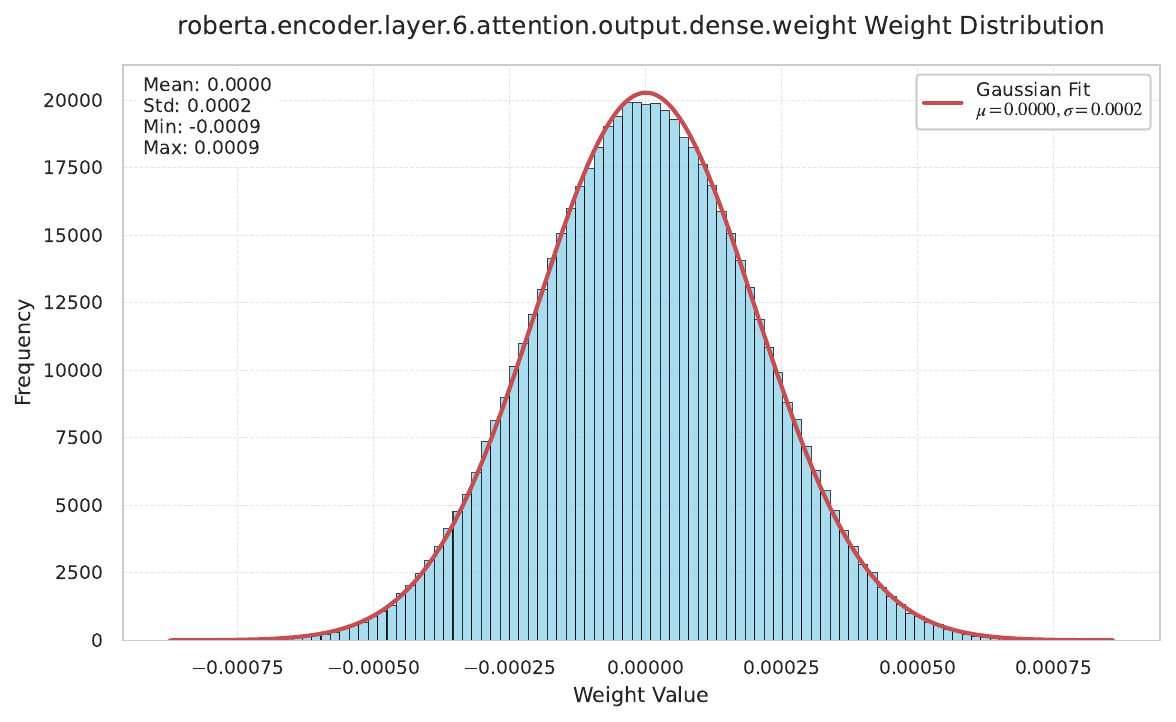}
        \caption{RoBERTa-Base (Attention Out Layer)}
        \label{fig:dist_roberta_attn_out}
    \end{subfigure}
    \hfill
    \begin{subfigure}[b]{0.33\textwidth}
        \centering
        \includegraphics[width=\linewidth]{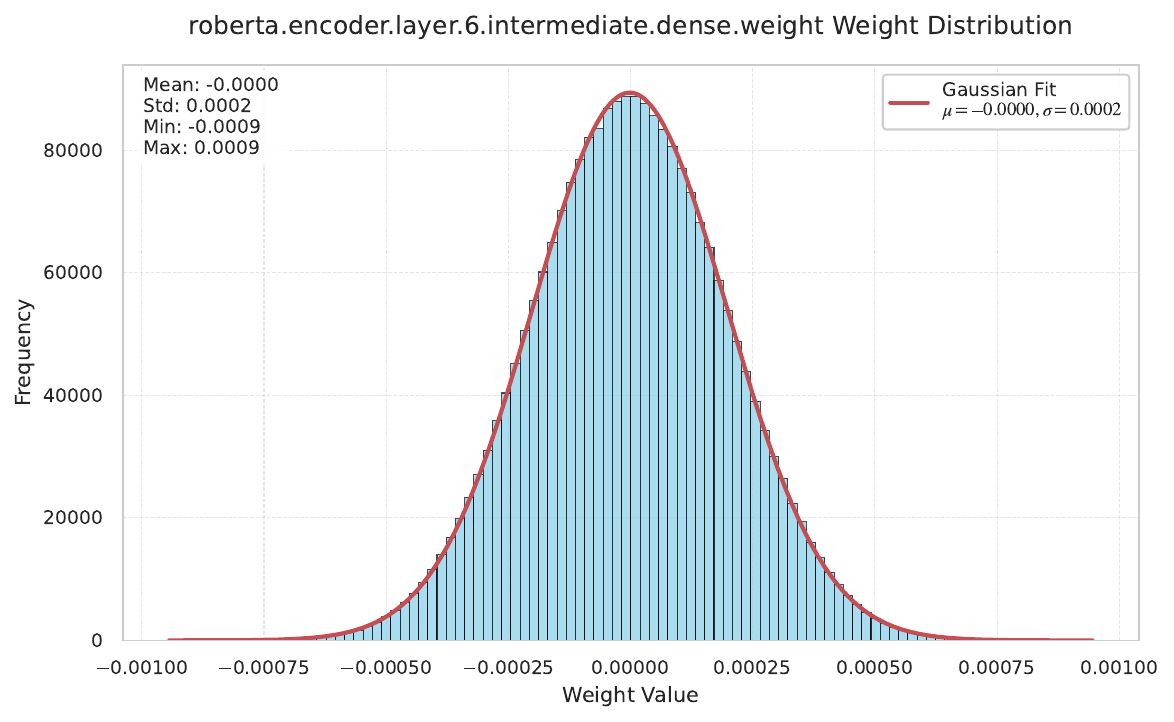}
        \caption{RoBERTa-Base (MLP Layer)}
        \label{fig:dist_roberta_mlp}
    \end{subfigure}
    \hfill
    \begin{subfigure}[b]{0.33\textwidth}
        \centering
        \includegraphics[width=\linewidth]{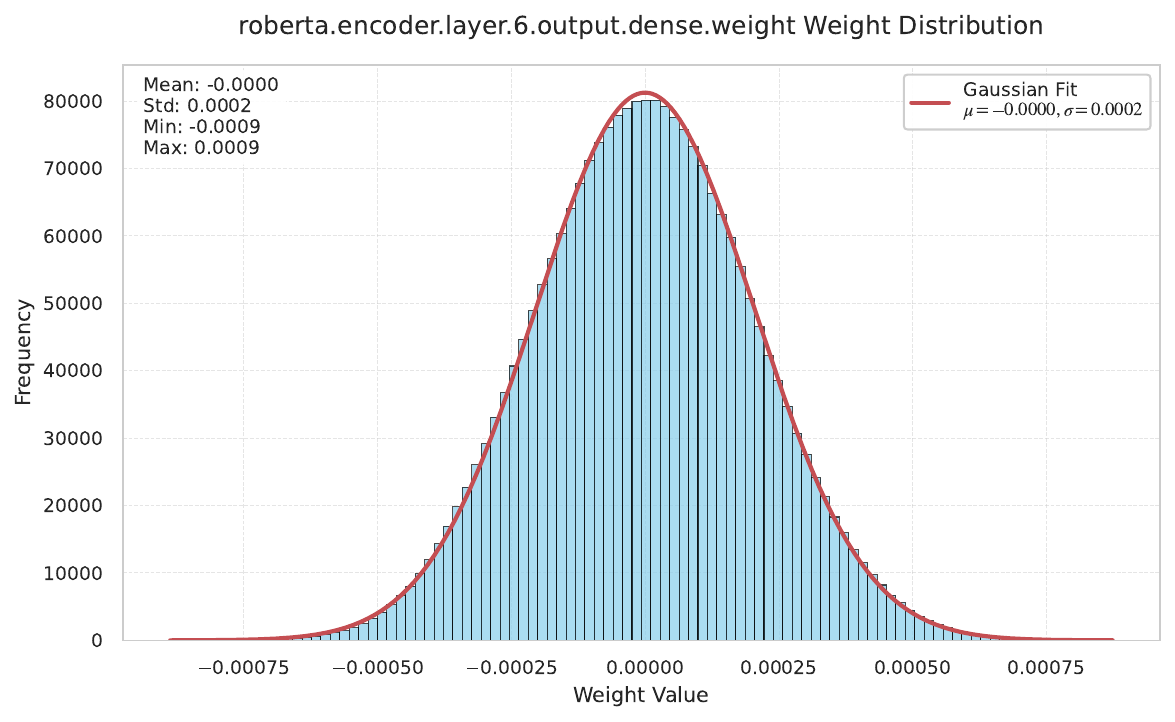}
        \caption{RoBERTa-Base (MLP Out Layer)}
        \label{fig:dist_roberta_mlp_out}
    \end{subfigure}

    \vspace{0.5cm}

    \begin{subfigure}[b]{0.33\textwidth}
        \centering
        \includegraphics[width=\linewidth]{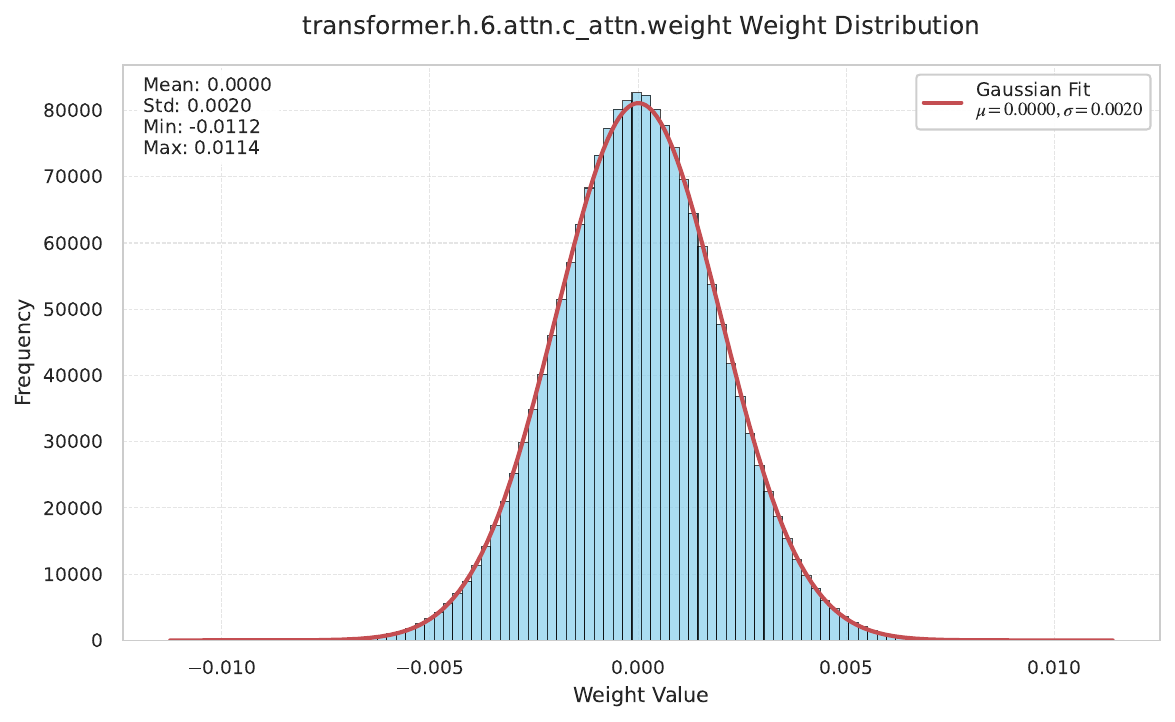}
        \caption{GPT-2 (Attention Layer)}
        \label{fig:dist_gpt2_attn}
    \end{subfigure}
    \hfill
    \begin{subfigure}[b]{0.33\textwidth}
        \centering
        \includegraphics[width=\linewidth]{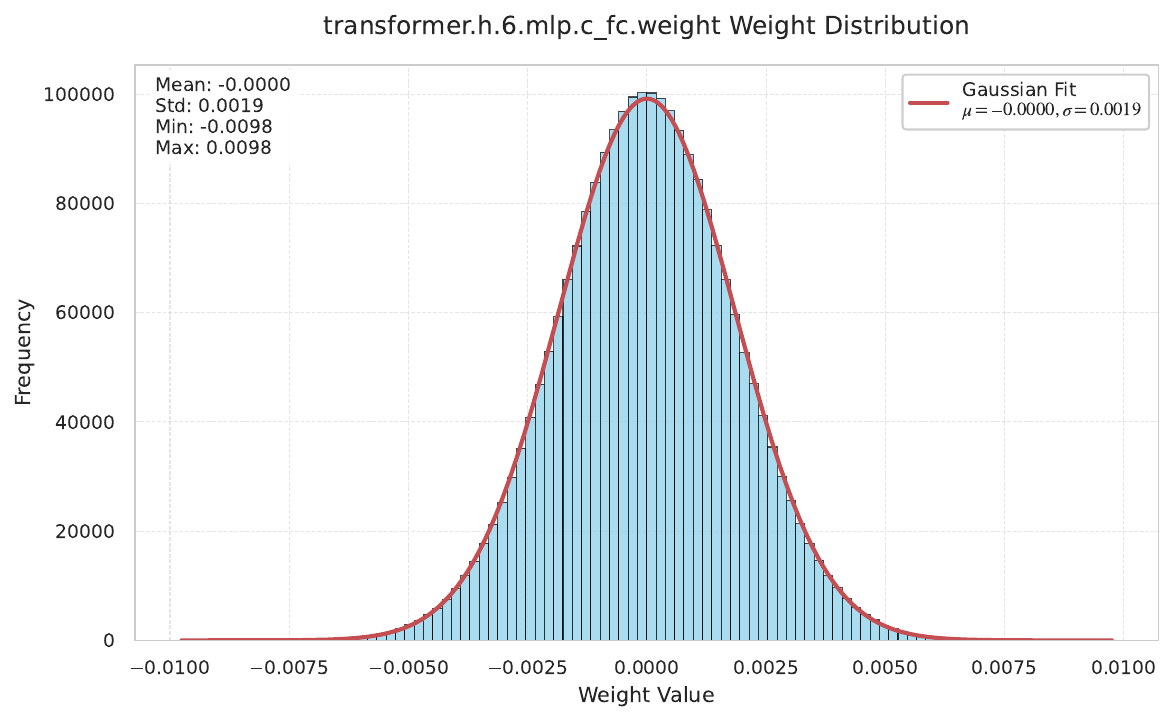}
        \caption{GPT-2 (MLP Layer)}
        \label{fig:dist_gpt2_mlp}
    \end{subfigure}
    \hfill
    \begin{subfigure}[b]{0.33\textwidth}
        \centering
        \includegraphics[width=\linewidth]{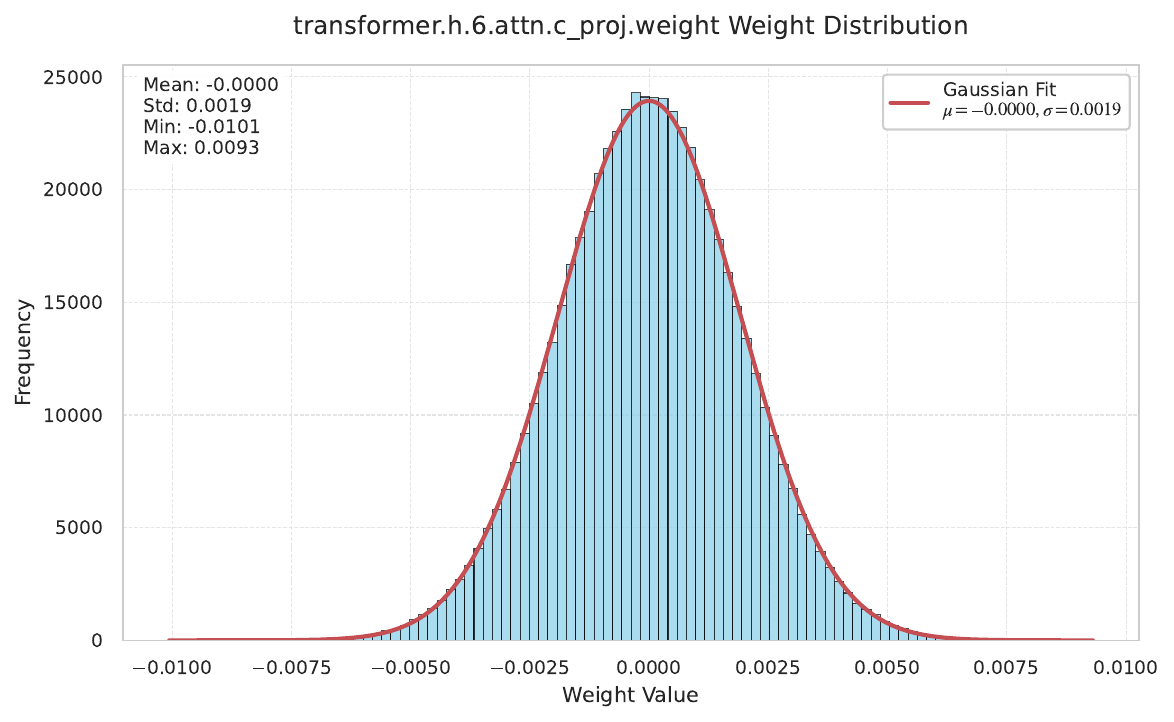}
        \caption{GPT-2 (Attention Proj Layer)}
        \label{fig:dist_gpt2_attn_proj}
    \end{subfigure}

    \vspace{0.5cm}

    \begin{subfigure}[b]{0.33\textwidth}
        \centering
        \includegraphics[width=\linewidth]{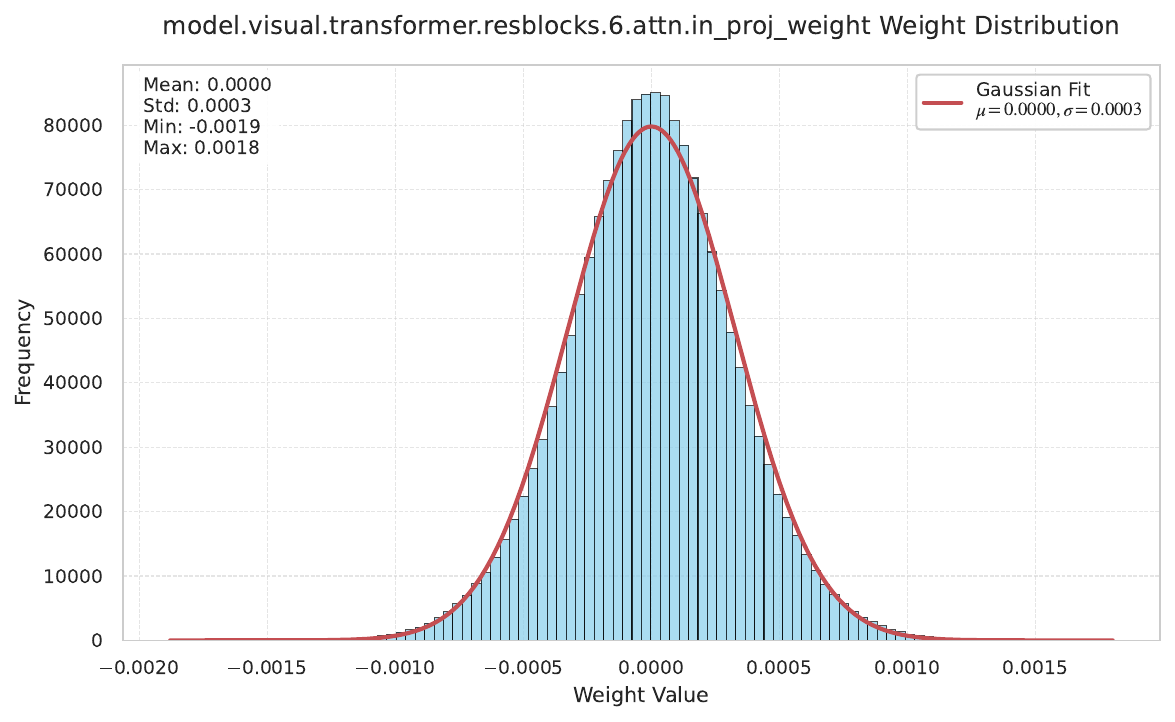}
        \caption{ViT-B/16 (Attention In Layer)}
        \label{fig:dist_vit_attn_in}
    \end{subfigure}
    \hfill
    \begin{subfigure}[b]{0.33\textwidth}
        \centering
        \includegraphics[width=\linewidth]{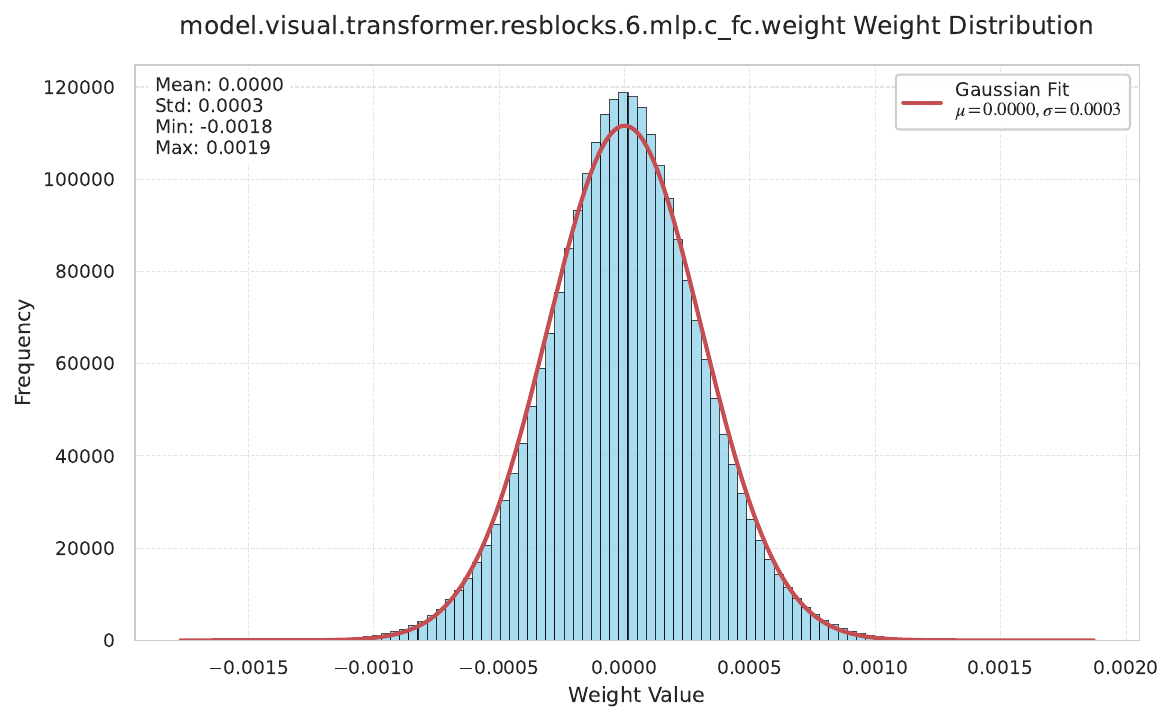}
        \caption{ViT-B/16 (MLP Layer)}
        \label{fig:dist_vit_mlp}
    \end{subfigure}
    \hfill
    \begin{subfigure}[b]{0.33\textwidth}
        \centering
        \includegraphics[width=\linewidth]{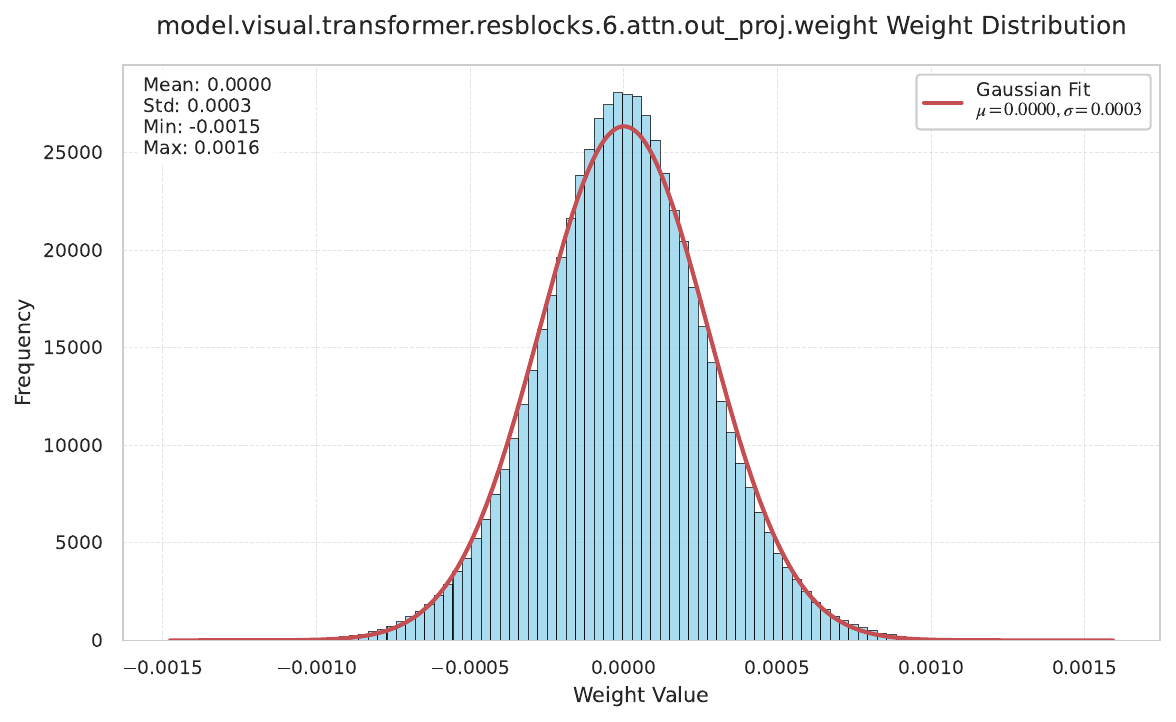}
        \caption{ViT-B/16 (Attention Out Layer)}
        \label{fig:dist_vit_attn_out}
    \end{subfigure}

    \caption[Empirical distributions of weight displacements]{
        \textbf{Empirical distributions of $\Delta W_t$ across architectures, layer types, and tasks.}
        Each subplot shows the histogram of the entries of $\Delta W_t$ for one representative layer from RoBERTa-Base, GPT-2, and ViT-B/16.
        The distributions are consistently zero-centered and approximately Gaussian, supporting the local zero-mean assumption used in our theoretical analysis and suggesting that the dominant task-specific information is captured by the second-order structure of the parameter updates.
    }
    \label{fig:app_delta_w_distributions}
\end{figure*}

\paragraph{Empirical validation of Assumption (A3).}
To validate Assumption~(A3), we examine the empirical distribution of the
entries of $\Delta W_t$ across architectures and layer types.
As shown in Fig.~\ref{fig:app_delta_w_distributions}, these distributions
are consistently centered around zero and exhibit approximately Gaussian shapes.

Under the linearized update model, a near-zero
mean distribution of the entries of $\Delta W_t$ is consistent with
$\mathbb{E}[e x^\top] \approx 0$.
This empirical observation supports Assumption~(A3) and suggests that the
dominant task-specific information is captured by the second-order structure
of the parameter updates.

\paragraph{Step 1: Linearized update expression.}
Under the local linearization in Assumption~(A1), the contribution of a single sample
to the first-order update is
\[
\Delta W^{(i)} = -2\eta\, e_i x_i^\top.
\]
Aggregating these first-order contributions over the fine-tuning trajectory gives
\begin{equation}
\Delta W_t \approx -2\eta \sum_{i=1}^{N_t} e_i x_i^\top.
\label{eq:app-deltaW}
\end{equation}

Let
\[
G_i = e_i x_i^\top, \qquad c = -2\eta,
\]
then
\[
\Delta W_t = c \sum_{i=1}^{N_t} G_i.
\]

\paragraph{Step 2: Second-order structure of the update.}
We analyze the column Gram matrix of the task vector:
\begin{align}
\Delta W_t^\top \Delta W_t
&=
c^2
\left(\sum_{i=1}^{N_t} G_i\right)^\top
\left(\sum_{j=1}^{N_t} G_j\right) \\
&=
c^2 \sum_{i,j} G_i^\top G_j.
\end{align}
Taking expectation gives
\begin{equation}
\mathbb{E}\!\left[\Delta W_t^\top \Delta W_t\right]
=
c^2 \sum_{i,j} \mathbb{E}[G_i^\top G_j].
\label{eq:app-gram-expand}
\end{equation}

Under Assumptions~(A2) and (A3), the cross terms vanish at leading order:
\[
\mathbb{E}[G_i^\top G_j] \approx 0,
\qquad i\neq j.
\]
Therefore, Eq.~\eqref{eq:app-gram-expand} reduces to
\begin{equation}
\mathbb{E}\!\left[\Delta W_t^\top \Delta W_t\right]
\approx
c^2 N_t\, \mathbb{E}[G^\top G],
\label{eq:app-reduction}
\end{equation}
where $G = e x^\top$ denotes a generic single-sample contribution.

\paragraph{Step 3: Single-sample second-order structure.}
For $G = e x^\top$, we have
\begin{align}
G^\top G
&=
(ex^\top)^\top (ex^\top) \\
&=
x e^\top e\, x^\top \\
&=
(e^\top e)\, xx^\top.
\end{align}
Taking expectation yields
\begin{align}
\mathbb{E}[G^\top G]
&=
\mathbb{E}\!\left[(e^\top e)\,xx^\top\right] \\
&\approx
\mathbb{E}[e^\top e]\,
\mathbb{E}[xx^\top]
\qquad \text{(A4)} \\
&=
\sigma_e^2 \Sigma_t,
\end{align}
where
\[
\sigma_e^2 = \mathbb{E}[e^\top e].
\]

Substituting into Eq.~\eqref{eq:app-reduction}, we obtain
\begin{equation}
\mathbb{E}\!\left[\Delta W_t^\top \Delta W_t\right]
\approx
c^2 N_t \sigma_e^2 \Sigma_t.
\end{equation}
Hence,
\begin{equation}
\mathbb{E}\!\left[\Delta W_t^\top \Delta W_t\right] \propto \Sigma_t,
\end{equation}
which proves Theorem~1.
\hfill$\square$

\paragraph{Remark on the practical estimator.}
In practice, only a single realization of $\Delta W_t$ is available for each task.
Therefore, rather than estimating the full population covariance of the fine-tuning trajectory,
we directly construct a stable second-order proxy from the observed task vector.

Following Eq.~(5) in the main text, we define the centered task vector
\[
\widetilde W_t
=
\Delta W_t - \mathbf{1}\mu_t^\top,
\qquad
\mu_t
=
\frac{1}{d_{\mathrm{out}}}\Delta W_t^\top \mathbf{1},
\]
where $\mathbf{1}\in\mathbb{R}^{d_{\mathrm{out}}}$ is the all-ones vector.
We then use the centered column Gram matrix
\[
\widehat{\Sigma}_t
\propto
\widetilde W_t^\top \widetilde W_t
\]
as the task-specific second-order proxy.

This centering removes the rank-one mean component across output dimensions,
so that $\widehat{\Sigma}_t$ better captures the task-specific second-order structure
emphasized by the update.
The proportionality constant affects only the overall scale and is handled later
by the adaptive normalization in Sec.~4.2.

\subsection{Additional statistics for adaptive covariance normalization}
\label{app:adaptive_cov_norm}

In practice, we compute the Frobenius norm via the trace identity
\[
\|\Delta W_t\|_F^2 = \mathrm{Tr}(\Delta W_t^\top \Delta W_t).
\]
Therefore, the heterogeneity coefficient introduced in
Sec.~\ref{sec:adaptive_cov_norm} can be implemented equivalently as
\begin{equation}
    \gamma
    =
    \frac{
        \mathrm{Var}_t\!\left[
            \log \mathrm{Tr}\!\left( \Delta W_t^\top \Delta W_t \right)
        \right]
    }{
        \left(
            \mathbb{E}_t\!\left[
                \log \mathrm{Tr}\!\left( \Delta W_t^\top \Delta W_t \right)
            \right]
        \right)^2
    }.
    \label{eq:gamma_trace_version}
\end{equation}
Since
\[
\|\Delta W_t\|_F^2 \equiv \mathrm{Tr}(\Delta W_t^\top \Delta W_t),
\]
Eq.~\eqref{eq:gamma_trace_version} is mathematically identical to the
original definition and aligns exactly with our implementation.

The coefficient $\gamma$ serves as a lightweight statistic for diagnosing
task-scale heterogeneity before adaptive normalization. Intuitively,
small $\gamma$ indicates that task updates have relatively aligned energy
scales, whereas large $\gamma$ reflects substantial variation across tasks
and signals stronger second-order mismatch.

In the main text, we reported results for RoBERTa-Base (8 tasks) and
ViT-B/16 (14 tasks), where $\gamma$ reliably separates relatively
homogeneous from highly heterogeneous task sets. To provide more
comprehensive evidence for the behavior of $\gamma$ across architectures
and task counts, we present additional statistics below.

\begin{figure*}[ht]
    \centering
    \begin{minipage}[b]{0.48\linewidth}
        \centering
        \includegraphics[width=\linewidth]{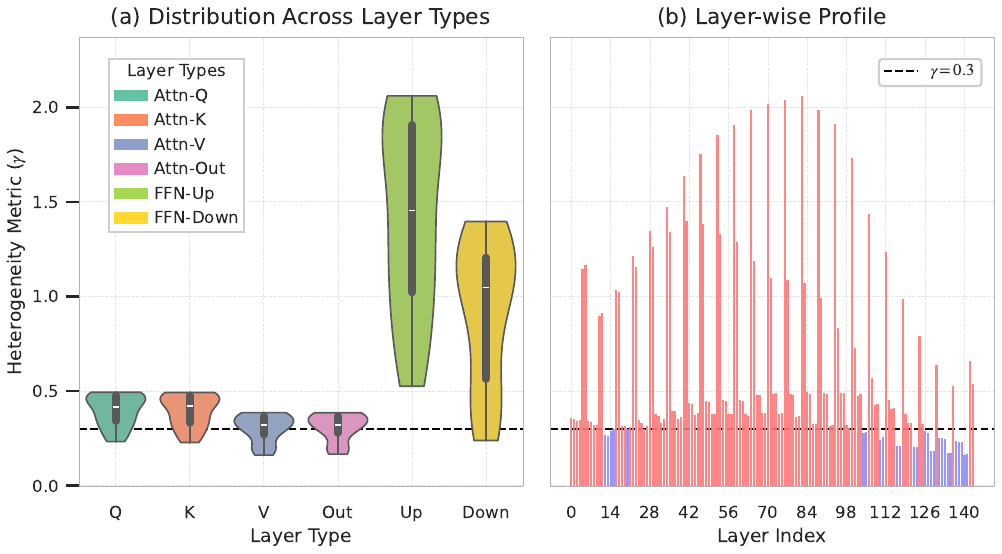}
        \caption{\textbf{RoBERTa-Large (8 tasks).}
        Trace statistics $\{\tau_t\}$ and heterogeneity coefficient~$\gamma$.}
        \label{fig:gamma_roberta_large}
    \end{minipage}
    \hfill
    \begin{minipage}[b]{0.48\linewidth}
        \centering
        \includegraphics[width=\linewidth]{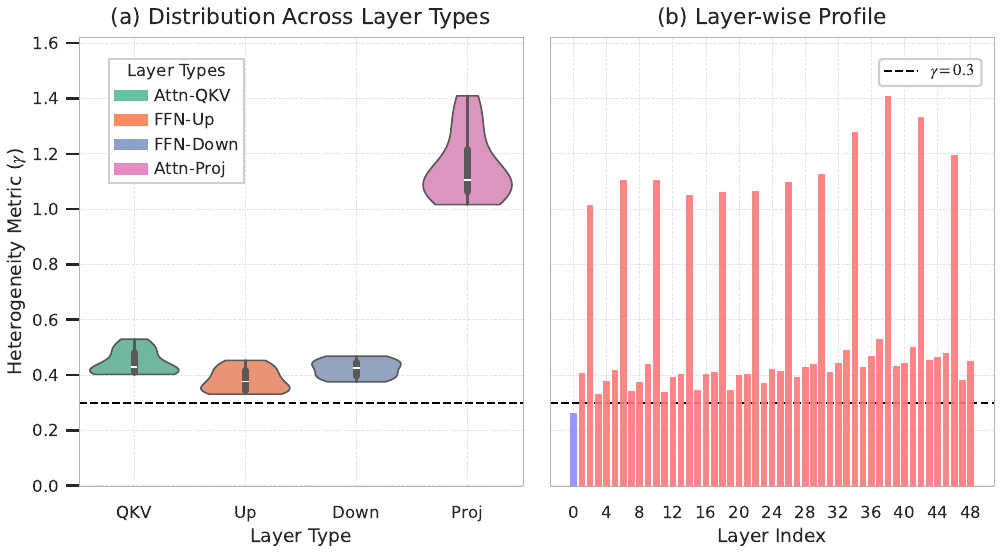}
        \caption{\textbf{GPT-2 (7 tasks).}
        Larger heterogeneity $\gamma$ than RoBERTa, indicating stronger
        mismatch across task updates.}
        \label{fig:gamma_gpt2}
    \end{minipage}
\end{figure*}

\paragraph{RoBERTa-Large on 8 tasks.}
Fig.~\ref{fig:gamma_roberta_large} shows the distribution of
$\{\tau_t\}$ and the resulting $\gamma$ for RoBERTa-Large across the same
8 GLUE-style tasks used in the main paper. Similar to RoBERTa-Base, the
heterogeneity score is high ($\gamma > 0.3$), confirming a substantial
degree of task specificity and scale mismatch across tasks.

\paragraph{GPT-2 on 7 tasks.}
Fig.~\ref{fig:gamma_gpt2} reports the corresponding values for GPT-2,
where we observe a noticeably larger spread in $\{\tau_t\}$ and a higher
$\gamma$ than for RoBERTa. This aligns with the empirical difficulty of
merging GPT-2 task vectors (Sec.~5), as greater heterogeneity makes simple
isotropic corrections less reliable and further motivates adaptive
normalization.

\begin{figure*}[h]
    \centering
    \begin{minipage}[b]{0.48\linewidth}
        \centering
        \includegraphics[width=\linewidth]{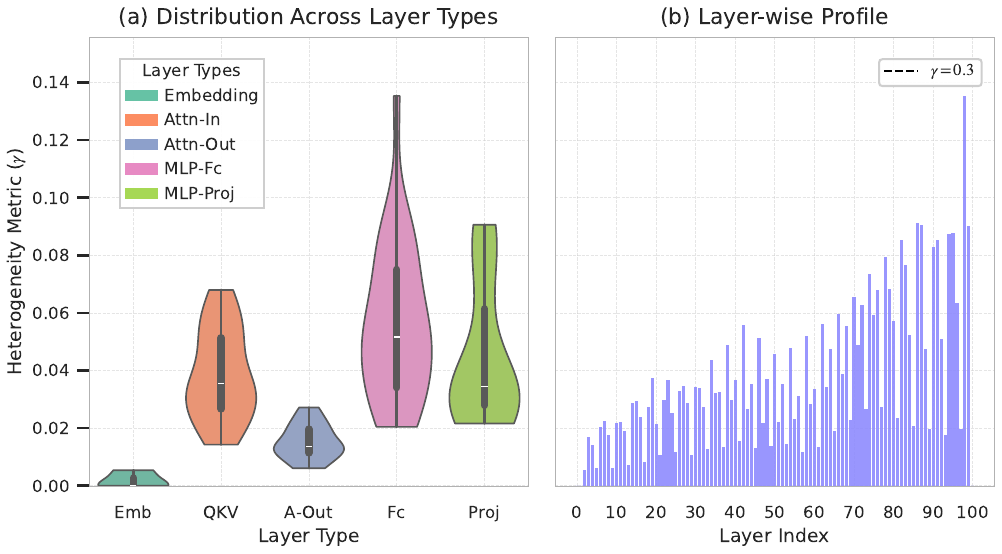}
        \caption{\textbf{ViT-L/14 (8 tasks).}
        Moderate heterogeneity $\gamma$, indicating manageable variation
        across task updates at smaller task scales.}
        \label{fig:gamma_vitl_8}
    \end{minipage}
    \hfill
    \begin{minipage}[b]{0.48\linewidth}
        \centering
        \includegraphics[width=\linewidth]{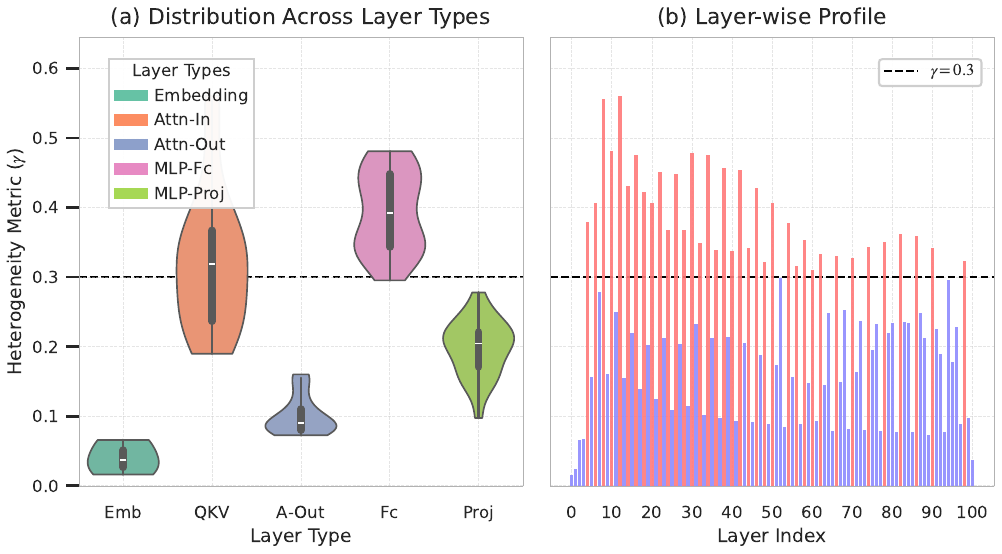}
        \caption{\textbf{ViT-L/14 (20 tasks).}
        Substantially larger heterogeneity $\gamma$, showing increasing
        mismatch across task updates as the number of merged tasks grows.}
        \label{fig:gamma_vitl_20}
    \end{minipage}
\end{figure*}

\paragraph{Effect of increasing number of tasks (ViT-L/14).}
To examine how heterogeneity evolves as more tasks are included, we
evaluate $\gamma$ for ViT-L/14 under multiple task-set sizes. In
Fig.~\ref{fig:gamma_vitl_8} and Fig.~\ref{fig:gamma_vitl_20}, we show the
results for 8 and 20 tasks, representing the low- and high-diversity
extremes. The heterogeneity coefficient increases noticeably when moving
from 8 to 20 tasks, consistent with the overall monotonic trend observed
across all task counts. This behavior supports our claim that task
diversity naturally grows as more datasets are merged, reinforcing the
need for the adaptive scaling mechanism introduced in
Sec.~\ref{sec:adaptive_cov_norm}.

\paragraph{Summary.}
Across all architectures, $\gamma$ consistently captures the heterogeneity
of the task set:
\begin{itemize}
    \item small $\gamma$ indicates relatively well-aligned task scales,
    for which simple merging procedures are often sufficient;
    \item large $\gamma$ reveals pronounced variability across tasks,
    motivating the adaptive covariance normalization used in \acem.
\end{itemize}
These results further support the robustness of $\gamma$ as a practical
criterion for activating heterogeneity-aware merging strategies.

\section{Computational Complexity and Practical Overhead}
\label{app:complexity}

We compare \acem\ with two representative data-free baselines:
the SVD-based TSV-M method~\cite{gargiulo2025tsvm}
and the gradient-based WUDI-merging approach~\cite{cheng2025whoever}.
In addition to asymptotic complexity, we also report practical
merge-time and peak GPU memory, following the quantitative
profiling added in the rebuttal.

We use the following notation.
Let $T$ be the number of task vectors and $L$ be the number of merged layers.
For a given layer, let each task update matrix $\Delta W_t$ have shape
$d_{\mathrm{out}} \times d_{\mathrm{in}}$.
For simplicity in big-$O$ notation, we define
\[
n = \max(d_{\mathrm{out}}, d_{\mathrm{in}})
\]
and report per-layer costs in terms of $n$.

\paragraph{TSV-M.}
TSV-M performs multiple singular value decompositions per layer:
\begin{itemize}
    \item $T$ SVDs on the task matrices $\Delta W_t \in \mathbb{R}^{d_{\mathrm{out}} \times d_{\mathrm{in}}}$,
    costing
    \[
    O\!\left(
        T \cdot \min(d_{\mathrm{out}}^2 d_{\mathrm{in}},\,
        d_{\mathrm{out}} d_{\mathrm{in}}^2)
    \right)
    = O(Tn^3).
    \]
    \item Two additional SVDs on aggregated matrices, contributing a lower-order cost of $O(n^3)$.
\end{itemize}
Therefore, the total per-layer complexity is
\[
O(Tn^3),
\]
and the overall complexity across all $L$ layers is
\begin{equation}
    O(\mathrm{TSV\mbox{-}M}) = O(LTn^3).
\end{equation}

\paragraph{Gradient-based merging (WUDI-merging).}
WUDI-merging optimizes the merged weights by iterative gradient descent.
Let $K$ denote the number of optimization steps.
Each step requires evaluating the loss and its gradient over all $T$ task vectors.
The dominant computation in each forward-backward pass consists of element-wise
operations on tensors of size
\[
T \times d_{\mathrm{out}} \times d_{\mathrm{in}},
\]
which costs
\[
O(T d_{\mathrm{out}} d_{\mathrm{in}})=O(Tn^2)
\]
per layer.
Hence the total complexity scales as
\begin{equation}
    O(\mathrm{WUDI}) = O(KLTn^2).
\end{equation}
Although the per-step cost is lower than cubic-time matrix decompositions,
the dependence on the optimization horizon $K$ makes WUDI computationally expensive
in practice, especially when hundreds or thousands of iterations are required.

\paragraph{ACE-Merging.}
\acem\ avoids iterative optimization and computes the merge in closed form.
Its per-layer complexity is dominated by three stages:
\begin{itemize}
    \item \textbf{Second-order proxy aggregation.}
    For each task, we form the centered column Gram proxy
    $\widetilde W_t^\top \widetilde W_t$
    (equivalently, up to centering, $\Delta W_t^\top \Delta W_t$),
    whose dominant matrix multiplication costs
    \[
    O(d_{\mathrm{out}} d_{\mathrm{in}}^2).
    \]
    Aggregating over all $T$ tasks gives
    \[
    O(T d_{\mathrm{out}} d_{\mathrm{in}}^2).
    \]

    \item \textbf{Closed-form solve.}
    After aggregation, the merged weights are obtained through a single
    inverse (or equivalently a linear solve) on a
    $d_{\mathrm{in}} \times d_{\mathrm{in}}$ system matrix, costing
    \[
    O(d_{\mathrm{in}}^3).
    \]

    \item \textbf{Spectral refinement (optional).}
    In the heterogeneous regime, ACE performs one additional SVD on a
    $d_{\mathrm{out}} \times d_{\mathrm{in}}$ matrix, costing
    \[
    O\!\left(
        \min(d_{\mathrm{out}}^2 d_{\mathrm{in}},\,
        d_{\mathrm{out}} d_{\mathrm{in}}^2)
    \right).
    \]
\end{itemize}

Combining these terms and using $n=\max(d_{\mathrm{out}}, d_{\mathrm{in}})$,
the per-layer complexity is
\[
O(Tn^3 + n^3) = O(Tn^3),
\]
so the total complexity across all $L$ layers is
\begin{equation}
    O(\acem) = O(LTn^3).
\end{equation}

\paragraph{Asymptotic comparison.}
ACE-Merging matches the asymptotic complexity of SVD-based methods such as TSV-M,
while retaining a true closed-form formulation.
Compared with gradient-based approaches such as WUDI-merging,
ACE avoids the additional multiplicative factor $K$ and is therefore
theoretically more scalable when many optimization steps are needed.

\paragraph{Practical overhead: time and peak memory.}
Beyond asymptotic complexity, we also profiled merge-time and peak GPU memory
on a single NVIDIA A800 (80GB), as reported in the rebuttal.
The results are summarized in Table~\ref{tab:complexity_runtime_memory}.
ACE-Merging is consistently the fastest method in both settings:
for ViT-L/14 (8 tasks), it reduces wall-clock time from
$126.68$s (WUDI, 1000 iterations) and $88.39$s (TSV-M)
to $4.32$s; for GPT-2 (7 tasks), it reduces merge time from
$62.71$s and $25.87$s to $4.70$s.
Its peak memory is also much lower than iterative baselines such as WUDI and Iso-CTS,
although TSV-M remains slightly more memory-efficient in these particular measurements.
These empirical results are consistent with the theoretical analysis above:
ACE is dominated by a one-pass second-order aggregation plus a single inversion/SVD,
whereas WUDI additionally scales with the optimization horizon $K$.%
\footnote{The table reports these measurements on a single A800 (80GB).}

\begin{table}[t]
\centering
\small
\caption{\textbf{Measured merge-time and peak GPU memory}
on a single A800 (80GB), reproduced from the rebuttal.}
\label{tab:complexity_runtime_memory}
\begin{tabular}{lcc|cc}
\toprule
& \multicolumn{2}{c|}{ViT-L/14 (8 tasks)} & \multicolumn{2}{c}{GPT-2 (7 tasks)} \\
Method & Time (s)$\downarrow$ & VRAM (GB)$\downarrow$ & Time (s)$\downarrow$ & VRAM (GB)$\downarrow$ \\
\midrule
WUDI (1000 iters) & 126.68 & 6.22 & 62.71 & 4.71 \\
TSV-M             & 88.39  & 1.91 & 25.87 & 0.77 \\
Iso-CTS           & 118.58 & 11.12 & 42.83 & 10.54 \\
ACE (ours)        & \textbf{4.32} & 2.04 & \textbf{4.70} & 1.02 \\
\bottomrule
\end{tabular}
\end{table}

\paragraph{Summary.}
Overall, the comparison suggests the following:
\begin{itemize}
    \item \textbf{vs.\ WUDI-merging:}
    ACE enjoys both a theoretical advantage
    (no dependence on the optimization horizon $K$)
    and a substantial practical speedup in measured wall-clock time.
    \item \textbf{vs.\ TSV-M:}
    ACE has the same asymptotic order $O(LTn^3)$,
    but in practice runs much faster in our experiments,
    while requiring only slightly higher peak memory.
\end{itemize}
Therefore, ACE-Merging offers a favorable trade-off between
closed-form stability, runtime efficiency, and practical scalability.

\subsection{Analysis of Limiting Cases and Hyperparameters}
\label{app:limiting_cases}

We analyze several limiting regimes of \acem\ to clarify how its main
hyperparameters affect the resulting merge.
These limiting cases should be interpreted as \emph{sanity checks} of the
formulation rather than practical operating regimes.
In particular, as emphasized in the rebuttal, the regularization strength
$\epsilon$ is introduced primarily for numerical stability of the closed-form
solve, while the default settings used throughout the experiments are fixed to
$\tau = 0.3$ and $k_{\mathrm{frac}} = 0.3$, with robustness observed under
moderate variations of these values.

\paragraph{Limit $\epsilon \to \infty$ (strong Tikhonov regularization).}
The hyperparameter \texttt{eps} ($\epsilon$) controls the strength of
Tikhonov regularization applied to each task-specific second-order proxy:
\[
\widehat{\Sigma}_t \leftarrow \widehat{\Sigma}_t + \epsilon_t I.
\]
As $\epsilon \to \infty$, the isotropic regularizer dominates the learned
second-order structure.
Two limiting behaviors arise depending on whether the task set is classified
as homogeneous or heterogeneous.

\begin{itemize}
    \item \textbf{Low heterogeneity ($\gamma \le \gamma_{\mathrm{thresh}}$).}
    In this regime, the same regularization coefficient is used for all tasks,
    i.e.\ $\epsilon_t=\epsilon$, so
    \[
    \widehat{\Sigma}_t \approx \epsilon I.
    \]
    Substituting into the closed-form merge gives
    \[
        \bar W
        =
        \left(\sum_{t=1}^T W_t \widehat{\Sigma}_t\right)
        \left(\sum_{t=1}^T \widehat{\Sigma}_t\right)^{-1}
        \xrightarrow[\epsilon\to\infty]{}
        \frac{1}{T}\sum_{t=1}^T W_t.
    \]
    Thus, in the infinitely isotropic limit, \acem\ reduces to
    \textbf{simple averaging} (task arithmetic).

    \item \textbf{High heterogeneity ($\gamma > \gamma_{\mathrm{thresh}}$).}
    In this regime, regularization is energy-adjusted across tasks, yielding
    \[
    \epsilon_t = \frac{\epsilon}{\tau_t},
    \qquad
    \tau_t = \mathrm{Tr}(W_t^\top W_t),
    \]
    so that
    \[
    \widehat{\Sigma}_t \approx \frac{\epsilon}{\tau_t} I.
    \]
    The merge then becomes
    \[
        \bar W
        \xrightarrow[\epsilon\to\infty]{}
        \frac{\sum_{t=1}^T \frac{1}{\tau_t} W_t}
             {\sum_{t=1}^T \frac{1}{\tau_t}},
    \]
    i.e.\ a \textbf{weighted average} in which tasks with larger Frobenius
    norms receive smaller effective weights.
\end{itemize}

These two limits help connect \acem\ to simpler merging rules, but they do
\emph{not} describe the practical regime used in our experiments.
As clarified in the rebuttal, $\epsilon$ is only a Tikhonov term for stable
inversion, and the degenerate limits discussed here are mainly included as
sanity checks; the default settings used in the paper are away from these
extremes.

\paragraph{Role of the heterogeneity threshold $\gamma_{\mathrm{thresh}}$.}
This threshold determines whether the method enters the
heterogeneity-aware branch, namely whether adaptive trace normalization and
spectral refinement are activated.

\begin{itemize}
    \item If $\gamma_{\mathrm{thresh}}$ is very large, the condition
    $\gamma > \gamma_{\mathrm{thresh}}$ is rarely satisfied.
    In that case, \acem\ remains in the low-heterogeneity branch and behaves
    like a regularized closed-form multi-task regression method without
    task-wise scale balancing.

    \item If $\gamma_{\mathrm{thresh}} \to 0$, the heterogeneous branch is
    always selected, so trace normalization and subsequent refinement are
    applied for essentially all task sets.
\end{itemize}

In practice, however, $\gamma_{\mathrm{thresh}}$ is not tuned per setting.
We use a fixed default threshold $\tau=0.3$ across both vision and language
experiments, and the sensitivity results show that performance remains stable
under moderate variation of this threshold.

\paragraph{Role of the spectral fraction $k_{\mathrm{frac}}$.}
This hyperparameter controls the rank of the spectral refinement applied after
the preliminary closed-form merge.
Let
\[
\bar W_{\mathrm{fused}} = U S V^\top,
\qquad
S = \mathrm{diag}(\sigma_1,\sigma_2,\ldots),
\]
and define the low-rank refinement
\[
\Delta W_{\mathrm{refine}}
=
\sigma_{\mathrm{iso}}\, U_{:,1:k} V_{:,1:k}^\top,
\qquad
\sigma_{\mathrm{iso}} = \frac{1}{k}\sum_{i=1}^k \sigma_i,
\]
where
\[
k = k_{\mathrm{frac}} \cdot \min(d_{\mathrm{in}}, d_{\mathrm{out}}).
\]

\begin{itemize}
    \item As $k_{\mathrm{frac}} \to 0$, we have $k \to 0$, so the refinement
    vanishes and the final merge reduces to the preliminary closed-form
    solution:
    \[
    \bar W \to \bar W_{\mathrm{pre}}.
    \]

    \item As $k_{\mathrm{frac}} \to 1$, the refinement spans the full singular
    basis, producing a full-rank spectral correction whose selected singular
    values are replaced by their mean.
    This yields a \emph{uniformized spectral correction} rather than a full
    isotropization of the merged weight.
\end{itemize}

Importantly, the purpose of $k_{\mathrm{frac}}$ is not to introduce an
arbitrary post-hoc modification, but to control how strongly the method
corrects the spectral collapse observed in the preliminary solution
$\bar W_{\mathrm{pre}}$.
As clarified in the rebuttal, the motivation is explicitly
\emph{pre-refinement}: Fig.~2 shows that $\bar W_{\mathrm{pre}}$ already
captures the correct dominant subspace, but exhibits an abnormally collapsed
spectrum relative to the expert models.
Spectral refinement is therefore a targeted correction that restores a
healthier spectral profile while preserving the dominant directions, rather
than a circular justification based on the final merged model.

Consistent with this interpretation, the rebuttal also notes that
$k_{\mathrm{frac}}=0.3$ is used as a fixed default across all experiments, and
that moderate changes in $k_{\mathrm{frac}}$ lead to only minor performance
variation.

\paragraph{Summary.}
These limiting regimes clarify how \acem\ interpolates between simpler and
more structured merging rules:
\begin{itemize}
    \item strong isotropic regularization recovers simple averaging or
    energy-weighted averaging;
    \item $\gamma_{\mathrm{thresh}}$ controls whether heterogeneity-aware
    normalization and refinement are activated;
    \item $k_{\mathrm{frac}}$ controls the strength of the targeted spectral
    correction applied to $\bar W_{\mathrm{pre}}$.
\end{itemize}
Together, these observations show that the hyperparameters of \acem\ have
clear functional interpretations, while the practical defaults remain fixed and
robust across architectures and modalities.

\section{ACE-Merging Performance on ViT Models with Increasing Task Coverage}
\label{app:vit_scaling}

We report the full performance of \acem\ on three vision transformer
backbones: ViT-B/32, ViT-B/16, and ViT-L/14, evaluated under increasing
numbers of downstream tasks. These results complement the heterogeneity
analysis in Sec.~\ref{sec:adaptive_cov_norm} and empirically demonstrate
that \acem\ remains stable as task diversity grows. All values are test
classification accuracies unless otherwise specified.

Each table presents side-by-side results for 8, 14, and 20 tasks, and
includes the average accuracy over all tasks for each backbone and
setting.

\begin{table*}[t]
\centering
\small
\begin{threeparttable}
\caption{\textbf{ACE-Merging accuracy (\%) on ViT-B/32, ViT-B/16, and ViT-L/14 across increasing numbers of tasks.}}
\label{tab:vit_all_scaling}
\setlength{\tabcolsep}{6pt} 

{\renewcommand{\arraystretch}{0.9} 
\begin{tabular}{@{}l@{\extracolsep{\fill}}ccc|ccc|ccc@{}}
\toprule
& \multicolumn{3}{c|}{\textbf{ViT-B/32}} 
& \multicolumn{3}{c|}{\textbf{ViT-B/16}} 
& \multicolumn{3}{c}{\textbf{ViT-L/14}} \\
\textbf{Dataset} 
& \textbf{8} & \textbf{14} & \textbf{20} 
& \textbf{8} & \textbf{14} & \textbf{20}
& \textbf{8} & \textbf{14} & \textbf{20} \\
\midrule
MNIST          & 99.46 & 99.22 & 92.65 & 99.44 & 99.32 & 96.40 & 99.58 & 99.59 & 96.65 \\
Cars           & 71.67 & 68.59 & 61.90 & 81.54 & 78.24 & 72.40 & 90.34 & 88.46 & 85.37 \\
DTD            & 89.63 & 82.45 & 75.80 & 90.69 & 83.19 & 76.33 & 96.97 & 93.19 & 88.67 \\
EuroSAT        & 95.81 & 88.37 & 83.70 & 97.78 & 96.96 & 94.56 & 99.52 & 99.30 & 98.22 \\
GTSRB          & 92.44 & 85.49 & 76.43 & 92.94 & 88.20 & 80.74 & 97.34 & 97.42 & 93.49 \\
RESISC45       & 86.79 & 81.51 & 77.92 & 90.73 & 88.75 & 85.97 & 94.76 & 93.25 & 92.11 \\
SUN397         & 73.06 & 70.75 & 68.63 & 76.57 & 74.04 & 72.07 & 81.83 & 78.62 & 76.72 \\
SVHN           & 94.19 & 90.10 & 82.89 & 95.08 & 93.22 & 88.40 & 96.20 & 95.61 & 91.80 \\
PCAM           & —     & 84.09 & 83.75 & —     & 86.14 & 85.02 & —     & 85.82 & 85.77 \\
CIFAR100       & —     & 72.77 & 68.79 & —     & 78.41 & 75.64 & —     & 86.64 & 83.75 \\
STL10          & —     & 97.50 & 96.73 & —     & 98.15 & 97.58 & —     & 99.41 & 99.34 \\
Oxford-IIIT Pet& —     & 90.54 & 88.20 & —     & 93.98 & 92.86 & —     & 96.43 & 96.18 \\
Flowers102     & —     & 73.04 & 70.19 & —     & 79.07 & 76.24 & —     & 87.32 & 84.57 \\
FER2013        & —     & 67.62 & 63.46 & —     & 68.04 & 63.81 & —     & 73.92 & 69.95 \\
CIFAR10        & —     & —     & 93.54 & —     & —     & 95.55 & —     & —     & 98.05 \\
Food101        & —     & —     & 80.71 & —     & —     & 88.47 & —     & —     & 93.98 \\
RenderedSST2   & —     & —     & 72.43 & —     & —     & 77.10 & —     & —     & 85.56 \\
EMNIST         & —     & —     & 98.07 & —     & —     & 97.82 & —     & —     & 99.52 \\
Fashion-MNIST  & —     & —     & 83.50 & —     & —     & 88.07 & —     & —     & 91.15 \\
KMNIST         & —     & —     & 57.27 & —     & —     & 37.23 & —     & —     & 78.95 \\
\midrule
\textbf{Average} 
& 87.88 & 82.29 & 78.83
& 90.60 & 86.12 & 82.11
& 94.57 & 91.07 & 89.49 \\
\bottomrule
\end{tabular}
} 
\end{threeparttable}
\end{table*}

\subsection{Additional results on LLaMA-3 and out-of-domain generalization}
\label{app:llama3}

We further evaluate \acem\ on a LLaMA-3 backbone fine-tuned on three
specialized expert task groups: multilingual QA (\texttt{multilingual}),
code generation (\texttt{coding}), and mathematical reasoning
(\texttt{math}). Each expert is fine-tuned on a disjoint subset of
benchmarks and then merged using different weight-space fusion
strategies. The merged models are evaluated jointly on five downstream
benchmarks: \texttt{xquad\_zh}, \texttt{xquad\_vi}, \texttt{gsm8k\_cot},
\texttt{mathqa}, and \texttt{humaneval}. The raw scores are reported in
Tab.~\ref{tab:llama3_raw}.

\begin{table}[t]
\centering
\small
\caption{\textbf{LLaMA-3 merging results (raw scores).}
Single-task fine-tuning experts and merged models evaluated on five
benchmarks.}
\label{tab:llama3_raw}
\setlength{\tabcolsep}{4pt}
{\renewcommand{\arraystretch}{0.9}
\begin{tabular}{lccccc}
\toprule
\textbf{Tasks} 
& \textbf{xquad\_zh}
& \textbf{xquad\_vi}
& \textbf{gsm8k\_cot}
& \textbf{mathqa}
& \textbf{humaneval} \\
\midrule
multilingual    & 27.24 & 42.20 &   --   &   --   &   --   \\
coding          &   --  &   --  &   --   & 34.30 & 49.39 \\
math            &   --  &   --  & 74.00  &   --  &   --   \\
\midrule
Average         &  8.40 & 32.72 & 73.62  & 34.24 & 46.34 \\
Task Arithmetic &  8.06 & 32.88 & 73.09  & 34.14 & 46.34 \\
ACE (ours)      & 16.42 & 37.68 & 69.75  & 34.57 & 50.61 \\
\bottomrule
\end{tabular}
}
\end{table}

To normalize for the different intrinsic difficulty and scale of each
benchmark, we further report in Tab.~\ref{tab:llama3_norm} the
performance of the merged models as a percentage of their corresponding
single-task fine-tuning expert. For each column, we divide the merged
score by the score of the relevant expert
(\texttt{multilingual} for \texttt{xquad\_zh} and \texttt{xquad\_vi},
\texttt{math} for \texttt{gsm8k\_cot}, and \texttt{coding} for
\texttt{mathqa} and \texttt{humaneval}) and multiply by~100.

\begin{table}[t]
\centering
\small
\caption{\textbf{Normalized performance on LLaMA-3 (\%).}
Each value is the ratio between the merged model and the best
single-task fine-tuning expert for that benchmark.}
\label{tab:llama3_norm}
\setlength{\tabcolsep}{4pt}
{\renewcommand{\arraystretch}{0.9}
\begin{tabular}{lcccccc}
\toprule
\textbf{Method} 
& \textbf{xquad\_zh}
& \textbf{xquad\_vi}
& \textbf{gsm8k\_cot}
& \textbf{mathqa}
& \textbf{humaneval}
& \textbf{Avg.} \\
\midrule
Average         & 30.84 & 77.54 & 99.49 & 99.83 & 93.82 & 80.30 \\
Task Arithmetic & 29.59 & 77.91 & 98.77 & 99.53 & 93.82 & 79.92 \\
ACE (ours)      & 60.28 & 89.29 & 94.26 & 100.79 & 102.47 & 89.42 \\
\bottomrule
\end{tabular}
}
\end{table}

The normalized results highlight the out-of-domain behavior of
\acem. On the multilingual benchmarks \texttt{xquad\_zh} and
\texttt{xquad\_vi}, which are out-of-domain for both the coding and math
experts, \acem\ recovers $60.28\%$ and $89.29\%$ of the corresponding
single-task fine-tuning performance, substantially outperforming both
simple averaging and task arithmetic (which remain around
$30$--$78\%$). On the code benchmark \texttt{humaneval}, \acem\ even
slightly exceeds the coding expert, achieving $102.47\%$ of the
fine-tuned score, while maintaining competitive accuracy on
\texttt{gsm8k\_cot} and \texttt{mathqa}. Overall, \acem\ attains the
highest normalized average score ($89.42\%$), indicating that it
effectively aggregates heterogeneous experts and exhibits strong
out-of-domain generalization across multilingual, coding, and
mathematical reasoning tasks.

\end{document}